\documentclass[10pt,twocolumn,letterpaper]{article}

\usepackage{iccv}
\usepackage{times}
\usepackage{epsfig}
\usepackage{graphicx}
\usepackage{amsmath}
\usepackage{amssymb}
\usepackage{multirow,tabulary,graphicx,subcaption,wrapfig}
\usepackage{booktabs}
\usepackage[english]{babel}

\usepackage[breaklinks=true,bookmarks=false]{hyperref}
\usepackage{url}

\iccvfinalcopy 

\def\iccvPaperID{2613} 

\ificcvfinal\fi

\begin{document}

\title{Cluster Alignment with a Teacher for Unsupervised Domain Adaptation}

\author{Zhijie Deng, Yucen Luo, Jun Zhu\thanks{Corresponding author.}\\
Dept. of Comp. Sci. \& Tech., Institute for AI, BNRist Lab, THBI Lab, Tsinghua University\\
{\tt\small \{dzj17, luoyc15\}@mails.tsinghua.edu.cn, dcszj@tsinghua.edu.cn}
}

\maketitle
\ificcvfinal\thispagestyle{empty}\fi

\begin{abstract}
Deep learning methods have shown promise in 
unsupervised domain adaptation, which aims to leverage a labeled source domain to learn a classifier for the unlabeled target domain with a different distribution. 
However, such methods typically learn a domain-invariant representation space to match the marginal distributions of the source and target domains, while ignoring their fine-level structures. In this paper, we propose Cluster Alignment with a Teacher (CAT) for unsupervised domain adaptation, which can effectively incorporate the discriminative clustering structures in both domains for better adaptation. Technically, CAT leverages an implicit ensembling teacher model to reliably discover the class-conditional structure in the feature space for the unlabeled target domain. Then CAT forces the features of both the source and the target domains to form discriminative class-conditional clusters and aligns the corresponding clusters across domains. Empirical results demonstrate that CAT achieves state-of-the-art results in several unsupervised domain adaptation scenarios. 
\end{abstract}


\section{Introduction}

\label{sec.intro}
Deep learning has achieved remarkable performance in a wide variety of computer vision tasks, such as image recognition~\cite{krizhevsky2012imagenet} and object detection~\cite{ren2015faster}. However, classifiers trained on specific datasets cannot always generalize effectively to new datasets owing to the well-known \emph{domain shift} problem~\cite{donahue2014decaf,torralba2011unbiased}. Enabling models to generalize from a source domain to a target domain is usually referred to as domain adaptation (DA)~\cite{ben2010theory}. 
In many cases, it is expensive or difficult to collect annotations on the target domain.
Learning algorithms attempting to tackle the transferring problem from a fully labeled source domain to an unlabeled target domain is called unsupervised domain adaptation (UDA)~\cite{gopalan2011domain}. UDA is particularly challenging because the target domain cannot provide explicit information to facilitate the adaptation of classifiers.

\begin{figure}[t]
\centering
  \vspace{-0.2cm}
  \includegraphics[width=0.9\linewidth]{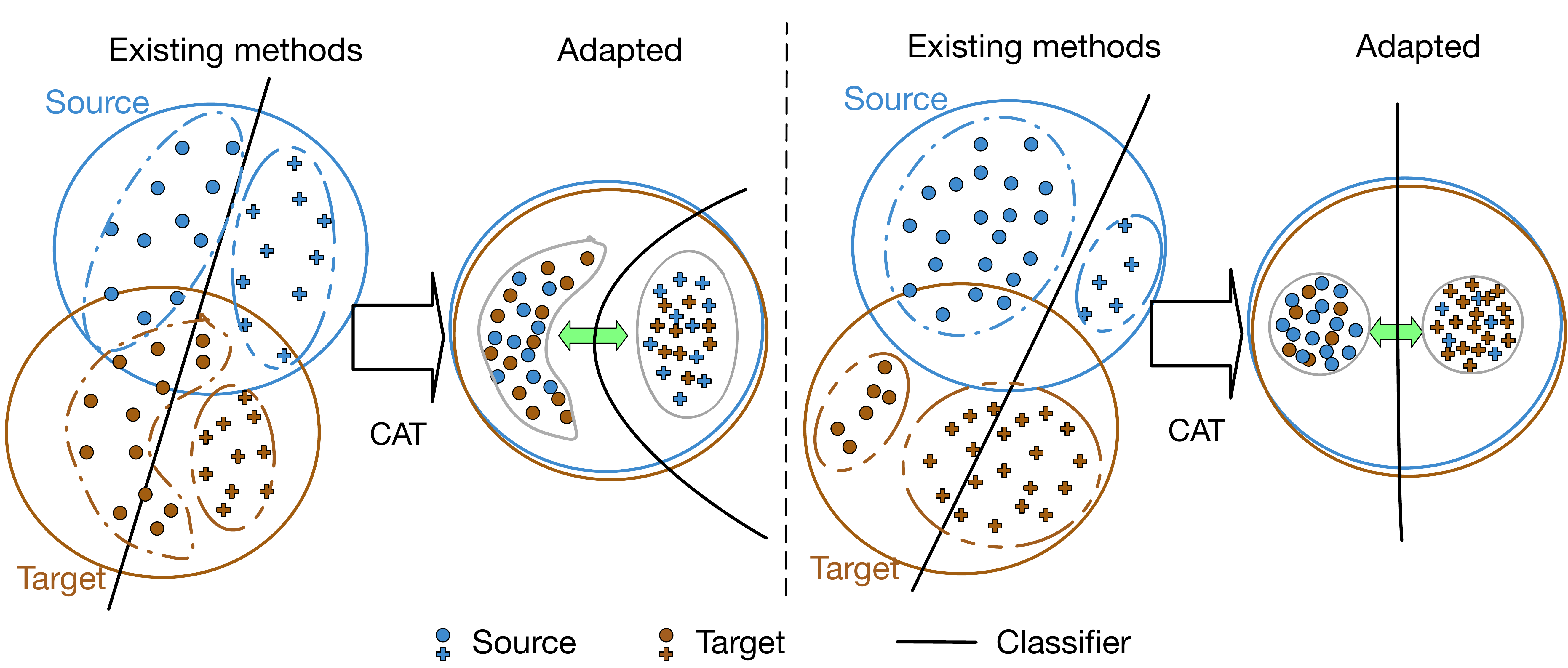}
    \vspace{-0.2cm}
\caption{(Best viewed in color.) Left: The two domains have diverse modes. Right: The two domains have different class imbalance ratios. Existing methods aligning the marginal distributions while ignoring the class-conditional structures cannot perform well in these cases. However, CAT incorporates the discriminative clustering structures in both domains for better adaptation, thus delivers a more reasonable domain-invariant cluster-structure feature space with enhanced discriminative power. See Fig.~\ref{fig:3} and Appendix. A for the learned feature space of real data.}  \vspace{-0.3cm}
\label{fig:0}
\end{figure}

Recently, deep models have been developed with promise in unsupervised domain adaptation to learn expressive features~\cite{tzeng2014deep,ganin2015unsupervised,long2015learning,liu2016coupled,tzeng2017adversarial,long2016deep,Saito_2018_CVPR,Sankaranarayanan_2018_CVPR,saito2017asymmetric}. These deep UDA methods mainly focus on matching the source and target domains via adversarial training~\cite{ganin2015unsupervised,tzeng2017adversarial,bousmalis2017unsupervised,liu2016coupled,long2016deep,xie2018learning,Saito_2018_CVPR,hoffman2017cycada} or kernelized training~\cite{long2015learning, long2016deep, long2016unsupervised}. 
The main hypothesis behind them is that
the marginal distributions of the two domains can be aligned in some feature space learned by optimizing a deep network, and thus the classifier trained with source data tends to perform well on the target domain. Theoretical analysis~\cite{ben2010theory} also shows that minimizing the divergence between the marginal distributions in the learned feature space is beneficial to reduce the classifier's error.


However, these methods are not problemless. 
In classification, as the classes correspond to different semantics and different characteristics, the marginal distribution of the data naturally has a class-conditional multi-modal structure. Moreover, the modes corresponding to the same class in different domains are not always geometrically similar. Thus, it is not sufficient for existing deep UDA methods to only minimize the discrepancy between the marginal distributions while neglecting their structures, and such methods tend to fail in challenging cases, such as those in Fig.~\ref{fig:0}. Properly incorporating this fine-grained class-conditional structure has been shown beneficial in various tasks. For example, Shi and Sha~\cite{shi2012information} 
make the \emph{discriminative clustering} assumption which 
helps to adapt the decision boundaries for the source domain to the target domain discriminatively.\footnote{Besides UDA tasks, previous work~\cite{pang2018max} has also shown an interesting exploration of the class-conditional structures for learning deep models that are robust against adversarial attacks.}  
However, one limitation of \cite{shi2012information} is that it adopts a simple linear transformation to learn the feature space, which cannot effectively extract high-order 
features from raw data (e.g., images) as the deep UDA methods. Another limitation is that  \cite{shi2012information} builds a nearest neighbor based prediction model, which outputs the prediction of one sample based on all the source data. Then, the training is not compatible with the stochastic training of deep network and has a high complexity.

In this paper, we present {\it Cluster Alignment  with  a  Teacher} (CAT), a new deep UDA model that 
incorporates the class-conditional structures for more effective adaptation. CAT conjoins the complimentary advantages of deep learning methods and discriminative clustering methods for UDA. 
Technically, there are three learning objectives in CAT. At first, CAT minimizes the supervised classification loss on the labeled source data and builds a teacher classifier, \ie an implicit ensemble of the source classifier, to provide pseudo labels for unlabeled target data. The underlying notion is that the golden classifier trained on source domain can perform well on a majority of target samples because of the similarity between the two domains and the teacher-student paradigm is not sensitive to the false pseudo labels~\cite{laine2016temporal}. To exploit the fine-grained class-conditional structures in the feature space and address the aforementioned issues suffered by existing deep UDA methods, CAT also includes two objectives which depend on the pseudo labels provided by the teacher classifier. On one hand, for discriminative learning in both domains, CAT deploys a class-conditional clustering loss to force the features from the same class to concentrate together and the features from different classes to be separated. On the other hand, for the class-conditional alignment between the two domains, CAT aligns the clusters which correspond to the same class but come from different domains via a conditional feature matching loss. The prediction models used in CAT are a student deep network and its implicit ensemble (\ie, the teacher classifier), thus CAT can address the training issues of \cite{shi2012information} and also enjoy the more flexible ability of feature learning. 
Furthermore, it is obvious that CAT is compatible to the marginal distribution alignment methods on the tasks where the source data is similarly distributed as the target data. The former can provide a fine-grained class-conditional alignment of domains and the latter can provide a global alignment of them.

We evaluate the proposed CAT through extensive experiments on both synthetic and real-world datasets.
Empirical results show that CAT presents striking performance across various tasks. In addition, we further combine CAT with the existing deep UDA methods, and CAT can bias them successfully to achieve the discriminative alignment between domains, establishing new state-of-the-art baselines on popular benchmarks. In the combined methods, we also propose a confidence-thresholding technique to filter out low-confidence target samples (which are likely to be mapped into incorrect clusters by the marginal distribution alignment methods) to enhance the stability of the training.

To summarize, our contributions  are three-folds:
\begin{itemize}
    \vspace{-0.2cm}
    \item We exploit the discriminative class-conditional structures of distributions in deep UDA and propose CAT to achieve better alignment between the source domain and the target domain.
    \vspace{-0.2cm}
    \item CAT is compatible and applicable to the existing UDA methods which rely on marginal distribution alignment.
    \vspace{-0.2cm}
    \item Empirically, CAT is not sensitive to hyper-parameters and can boost the marginal distribution alignment approaches significantly, achieving new state-of-the-art across various settings.
\end{itemize}

\vspace{-0.3cm}
\section{Related work}
\vspace{-0.1cm}
Unsupervised domain adaptation has drawn increasing interests
, and has been developed mainly in two directions: Maximum Mean Discrepancy (MMD) based approaches and adversarial training based approaches. Tzeng \etal~\cite{tzeng2014deep} and Long \etal~\cite{long2015learning} minimize MMD to match the two domains while~\cite{long2016deep} proposes to align the joint distributions of them using Joint MMD criterion. Since the development of Generative Adversarial Networks (GANs)~\cite{goodfellow2014generative,deng2017structured}, adversarial training has been applied into domain adaptation and fruitful works emerge. Ganin \etal~\cite{ganin2015unsupervised} develop the framework of domain adversarial training and plenty of works are proposed to improve it by aligning source domain and target domain better in the feature space~\cite{tzeng2017adversarial,xie2018learning,hong2018conditional} or image space~\cite{liu2016coupled}. Zhang \etal~\cite{zhang2018collaborative} successfully extend RevGrad~\cite{ganin2015unsupervised} to consider each domain's characteristics using collaborative games. Image to image translation approaches~\cite{ghifary2016deep,bousmalis2017unsupervised,hoffman2017cycada,russo2017source,murez2017image,Sankaranarayanan_2018_CVPR} also play an important role in the advancement of domain adaptation and demonstrate impressive performance, especially on semantic segmentation tasks. 
In addition, Saito \etal~\cite{Saito_2018_CVPR} propose to align the two domains using decision boundaries of task-specific classifier. Associative DA~\cite{haeusser2017associative} proposes an associative loss to reduce discrepancy between domains and SimNet~\cite{pinheiro2018unsupervised} proposes to use a similarity-based classifier in UDA. Though the existing methods match the two domains in different ways, most of them ignore the class-conditional information in the alignment procedure, thus hard to attain the objective of discriminative learning. Conversely, CAT explicitly discovers classes in the feature space via a teacher model and hence constructs a more reasonable matching procedure. Concurrently, several  works~\cite{zhao2019learning,wu2019domain} analyze the conditional distribution shift and label distribution shift issues in UDA theoretically, but only propose limited solutions. In contrast, we provide a more practical and more powerful way to solve them.

Using a teacher model for labeling data is inspired by the impressive consistency-based methods in semi-supervised learning (SSL)~\cite{laine2016temporal,tarvainen2017mean}. 
Recent attempts to apply SSL techniques in UDA include~\cite{french2017self,shu2018dirt,volpi2017adversarial}. CAT differs from these previous works in that CAT exploits the discriminative class-conditional structures in both the alignment and classification procedures
while they focus on improving the classifier for the target domain by implementing the \emph{cluster} assumption~\cite{chapelle2009semi}.
CAT imposes a much stronger regularization and assists in a better alignment.

\vspace{-0.1cm}
\section{Methodology}
\label{sec.main}
\vspace{-0.1cm}
In this section, we first introduce the setting and framework of deep UDA and then present the \emph{Cluster Alignment with a Teacher} (CAT). 
Finally, we discuss about CAT.

\vspace{-0.1cm}
\subsection{Deep unsupervised domain adaptation}
\vspace{-0.1cm}
In an UDA task, we are given a set of source samples $\mathcal{X}_s=\{x_s^i\}_{i=1}^N$ with labels $\mathcal{Y}_s=\{y_s^i\}_{i=1}^N$, $y_s^i\in \{1,2,...,K\}$ and a set of unlabeled target samples $\mathcal{X}_t=\{x_t^i\}_{i=1}^M$. 
Notably, the two sets of samples are drawn from different distributions which lead to the \emph{domain shift} challenge. Therefore, the UDA algorithms should learn to adapt the classifier trained on the source domain to the unlabeled target domain. 
Deep learning techniques have been introduced into UDA~\cite{ganin2015unsupervised,tzeng2017adversarial,bousmalis2017unsupervised,liu2016coupled,long2016deep,Saito_2018_CVPR,long2015learning,long2016unsupervised} and they demonstrate remarkable performance across tasks. Generally, in these methods, the classifier $h$ (parameterized by $\theta$) is constructed as $h=g\circ f$ where $f$ maps samples into features in the space $\mathcal{F}$ and $g$ outputs the predictions based on the extracted features. The learning includes simultaneously optimizing the classifier $h$ w.r.t. the labeled source data and minimizing the distance between the marginal distributions of the two domains in the feature space $\mathcal{F}$, resulting in a domain-invariant feature space. Technically, in the source domain, we minimize the supervised loss as:
\vspace{-0.1cm}
\begin{equation}\label{eq-1}
    \min_{\theta} \mathcal{L}_{y}(\mathcal{X}_s, \mathcal{Y}_s) = \frac{1}{N}\sum_{i=1}^N \ell(h(x_s^i;\theta), y_s^i),
\end{equation}
where $\ell$ is a pre-defined loss, \eg, cross-entropy loss. 
Meanwhile, we minimize the discrepancy loss as:
\begin{equation}
\label{eq-1.2}
    \min_{\theta} \mathcal{L}_{d}(\mathcal{X}_s, \mathcal{X}_t) =  
    \mathcal{D}(f(\mathcal{X}_s, \theta), f(\mathcal{X}_t, \theta)),
\end{equation}
where $\mathcal{D}$ is a distance and usually correlated with the ${\mathcal{H}\Delta\mathcal{H}}$ distance in the error bound theory of DA~\cite{ben2010theory}. The theory reveals that the expected error on target samples of any classifier $h$ drawn from a hypothesis set $\mathcal{H}$ has the following bound~\cite{ben2010theory,xie2018learning}:
\begin{equation}
\label{eq-10}
\begin{split}
    \epsilon_{t}(h) & \leq \epsilon_{s}(h) + \frac{1}{2}d_{\mathcal{H}\Delta\mathcal{H}}(s,t) \\
    & \;\;\;\;\;\;\;\;\;\;\;\;\; + \min_{\hat{h}\in\mathcal{H}}(\epsilon_{s}(\hat{h},l_{s})+\epsilon_{t}(\hat{h},l_{t})) \\
    & \leq \epsilon_{s}(h) + \frac{1}{2}d_{\mathcal{H}\Delta\mathcal{H}}(s,t) + \epsilon_{t}(l_{s},l_{t}) \\
    & \;\;\;\;\;\;\;\;\;\;\;\;\; + \min_{\hat{h}\in\mathcal{H}}(\epsilon_{s}(\hat{h},l_{s})+\epsilon_{t}(\hat{h},l_{s})), \\
\end{split}
\end{equation}
where $\epsilon_{s}(h)$ denotes the expected error on source samples of $h$, and $l_{s}$ and $l_{t}$ represent the labelling functions~\cite{ben2010theory} for the source and target domains, respectively. $\epsilon_{t}(l_{s},l_{t})$ denotes the disagreement between the labelling functions in the target domain. Notably, $\min_{\hat{h}\in\mathcal{H}} (\epsilon_{s}(\hat{h},l_{s})+\epsilon_{t}(\hat{h},l_{s}))$ can be small enough by optimizing $\hat{h}$ w.r.t. the labeled source data. 
The supervised loss and discrepancy loss focus on minimizing $\epsilon_{s}(h)$ and $d_{\mathcal{H}\Delta\mathcal{H}}(s,t)$ respectively to obtain small target domain classification error.

However, the methods working in the above framework are not problemless. Theoretically, they ignore minimizing $\epsilon_{t}(l_{s},l_{t})$ which may lead to a large upper bound of $\epsilon_{t}(h)$ and result in unsatisfying target domain performance~\cite{xie2018learning}. Empirically, the data in classification naturally has a class-conditional multi-modal structure, thus aligning marginal distributions while ignoring the fine-level discriminative structures of domains may hurt the target classification performance (Fig.~\ref{fig:0}-left). Moreover, these methods may fail in more practical and challenging problems, \eg the source and target domains have obviously different class imbalance ratios (Fig.~\ref{fig:0}-right).



\subsection{Cluster Alignment with a Teacher}
\begin{figure}[t]
\centering
  \vspace{-0.3cm}
  \includegraphics[width=\linewidth]{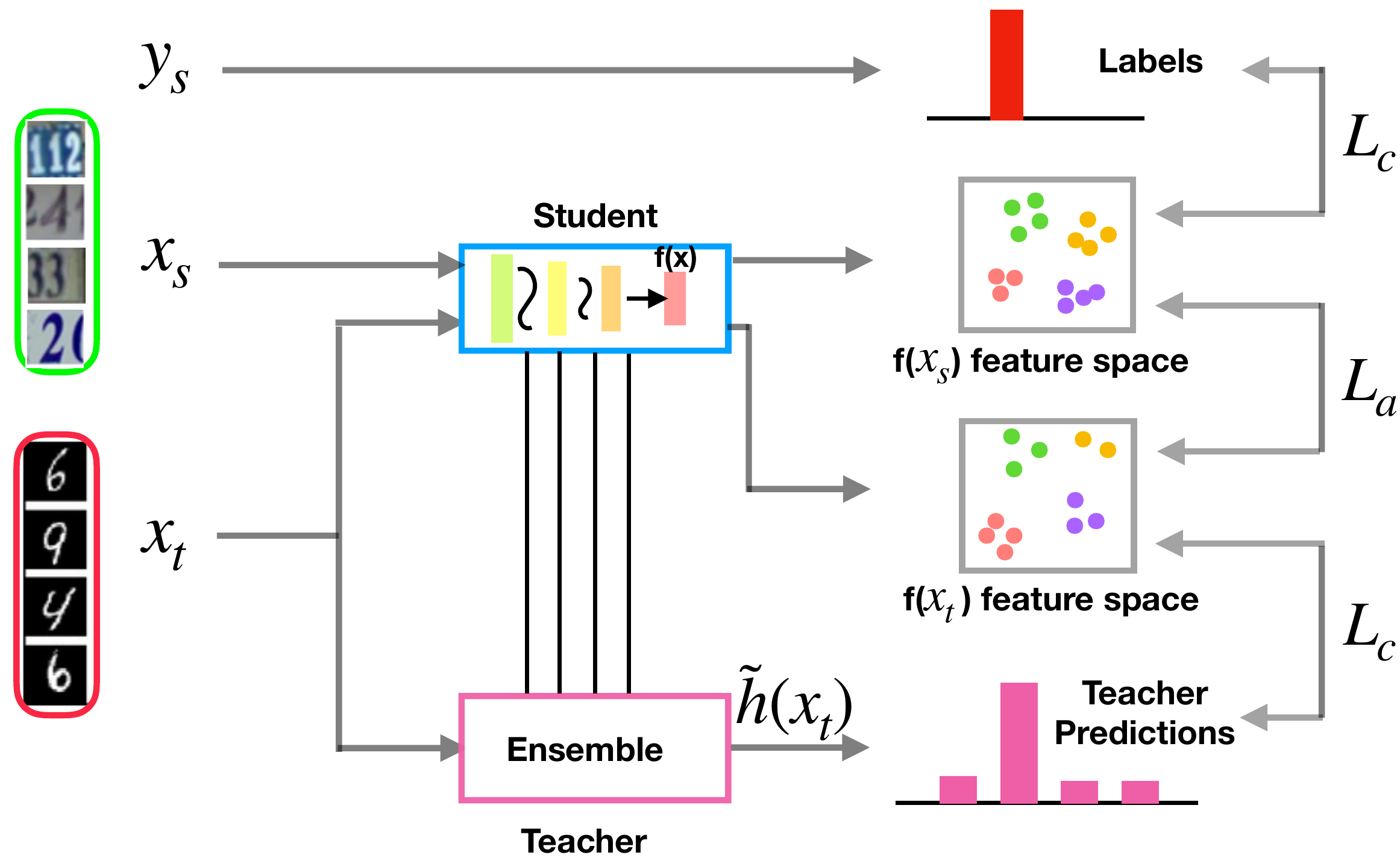}
  \vspace{-0.8cm}
\caption{The framework of CAT (The source supervised loss $\mathcal{L}_y$ is omitted for clarity).}  \vspace{-0.4cm}
\label{fig:framework}
\end{figure}

To overcome these issues, we expect to exploit the fine-level structures in the feature space for discriminative learning and match the class-conditional distributions of source and target domains to reduce of the mismatching between $l_{s}$ and $l_{t}$. Therefore, in the deep UDA scenario, we propose \emph{Cluster Alignment with a Teacher} (CAT), a new deep UDA model for more effective adaptation. Specifically, for the objectives of discriminative learning and class-conditional alignment between domains, we propose a discriminative clustering loss $\mathcal{L}_c$ to force the features of both the source and the target domains to form discriminative clusters, and a cluster-based alignment loss $\mathcal{L}_a$ to align the clusters corresponding to the same class in different domains. Given them, we propose to train CAT 
by solving the following optimization problem:
\begin{equation}
    \min_{\theta}\mathcal{L}_y + \alpha( \mathcal{L}_{c} + \mathcal{L}_{a}),
\end{equation}
where the hyper-parameter $\alpha$ sets a relative trade-off. 
The whole framework is shown in Fig.~\ref{fig:framework}. We build a teacher classifier, \ie an implicit ensemble of the classifier to be optimized, to provide pseudo labels for the unlabeled target data. These pseudo labels will be used in $\mathcal{L}_c$ and $\mathcal{L}_a$.
We use stochastically sampled mini-batches in the two objectives and the two classifiers make predictions in a forward-propagation way, thus CAT can be trained more efficiently than \cite{shi2012information}. Furthermore, $\mathcal{L}_c$ and $\mathcal{L}_a$ optimize the feature space directly and will be more effective than the nearest neighbor based clustering loss in \cite{shi2012information}. We elaborate $\mathcal{L}_c$ and $\mathcal{L}_a$ in the following sections.
\subsubsection{Discriminative clustering with a teacher}

For better classification and alignment, we propose to discover the class-conditional structures in the feature space in both the source and the target domains, and then shape them to be discriminative clusters.
In the source domain, the class-conditional structure is obvious because the data is fully labeled. Nevertheless, in the target domain, we cannot obtain the class-conditional structure easily due to the lack of labels. The semantic similarity between the two domains implies that the classifier $h$ trained on the source domain can predict most of target samples correctly. Consequently, using pseudo labels~\cite{lee2013pseudo} as the annotations for target data and conducting self-training is a direct approach, but it suffers from the error amplification issue which can be detrimental to the learning procedure. To discover the class-conditional structure of the target features in a reliable way, we introduce a teacher classifier $\tilde h$ defined as an implicit ensemble of the previous student classifier $h$~\cite{laine2016temporal} to provide pseudo labels for target data.

Based on the 
pseudo labels given by the teacher, we can explicitly force the target class-conditional structure to be more discriminative using a clustering loss. 
For the source domain, a similar one can be applied. Formally, resembling the effective SNTG loss in \cite{luo2017smooth}, we employ the following discriminative clustering loss (we omit the dependence of $f$ on $\theta$ for clarity, unless stated otherwise):
\begin{equation}
\begin{split}
    \mathcal{L}_{c}(\mathcal{X}_s,\mathcal{X}_t) = \mathcal{L}_{c}(\mathcal{X}_s) + \mathcal{L}_{c}(\mathcal{X}_t) ,
\end{split}
\end{equation}
\begin{equation}
\begin{split}
    \mathcal{L}_{c}(\mathcal{X})&=\frac{1}{|\mathcal{X}|^2}\sum_{i=1}^{|\mathcal{X}|}\sum_{j=1}^{|\mathcal{X}|}\left[\delta_{ij}d\left(f(x^i), f(x^j)\right) + \right.\\ 
    &\left.(1-\delta_{ij})\max\left(0,m-d\left(f(x^i), f(x^j)\right)\right)\right], \\
\end{split}
\end{equation}
where $d$ is the distance (\eg, squared Euclidean distance) between two features, $m$ is a pre-defined margin, and $\delta_{ij}$ is an indicator function which outputs 1 only if $x^i$ and $x^j$ have the same ground truth label (source domain) or teacher-annotated label (target domain). $\mathcal{L}_c$ encourages the features from the same class to concentrate together and pushes the features from different classes far away from each other with a distance $m$ at least. This loss modifies 
the structures in the representation space gradually, and consequently, it demonstrates a class-conditional cluster structure (as shown in Sec.~\ref{sec-exp4}). Note that minimizing $\mathcal{L}_c$ is consistent with the \emph{cluster} assumption~\cite{luo2017smooth} of classifier and benefits the performance of classification.

It's a common doubt whether the incorrect predictions of the teacher classifier would destroy the training dynamics. However, previous works on semi-supervised learning~\cite{laine2016temporal, tarvainen2017mean} have validated that this kind of training always leads to good convergence and demonstrates robustness against incorrect labels. Intuitively, the teacher instructs the training of one instance through a bundle of others' predictions which alleviates the negative influence of incorrect predictions notably and aids the student to give better predictions.

\subsubsection{Cluster alignment via conditional feature matching}
Once the feature space presents discriminative cluster structure, the classifier is expected to make more accurate predictions. However, the label predictor $g$ trained on source domain features may fail due to the geometrical mismatching between the clusters which correspond to the same class in different domains. This kind of mismatching is brought by the individual characteristics of each domain. As a result, it is necessary to impose 
a class-conditional alignment of two domains to learn better domain-invariant features and adjust the target feature space to be more suitable for classification. Naturally, we expect to minimize the divergence between the corresponding clusters in source domain and the teacher-annotated target domain:
\begin{equation}
    \min_{\theta} \mathcal{D}(\mathcal{F}_{s,k} || \mathcal{F}_{t,k}),
\end{equation}
where $\mathcal{F}_{s,k}$ ($\mathcal{F}_{t,k}$) denotes the set consisting of all the features belonging to class $k$ of domain $s$ (domain $t$).
Extensive previous works~\cite{goodfellow2014generative,li2015generative} have been proposed to minimize the distance between two sets of samples but we expect to achieve this in a more simple and efficient way by exploiting the separable and tight clusters in the feature space. Drawing inspiration from feature matching GANs~\cite{salimans2016improved} which optimizes the distance between the first-order statistics of distributions and demonstrates striking results on SSL tasks, we choose to extend it to work in a conditional way. Formally, we introduce the following cluster alignment loss:
\begin{equation}\label{eq:(6)}
    \begin{split}
    \mathcal{L}_{a}(\mathcal{X}_s, \mathcal{X}_t) &=  
    \frac{1}{K}\sum_{k=1}^K\parallel\lambda_{s,k}-\lambda_{t,k}\parallel^2_2,
    \end{split}
\end{equation}
where $\lambda_{s,k}$ and $\lambda_{t,k}$ are calculated by
\begin{equation}\label{eq:6}
    \lambda_{s,k} = \frac{1}{|\mathcal{X}_{s,k}|}\sum_{x_{s}^i\in\mathcal{X}_{s,k}}f(x_{s}^i), \,
    \lambda_{t,k} = \frac{1}{|\mathcal{X}_{t,k}|}\sum_{x_{t}^i\in\mathcal{X}_{t,k}}f(x_{t}^i)
\end{equation}
where $\mathcal{X}_{s,k}$ is the subset of $\mathcal{X}_{s}$ containing all the source samples whose ground-truth labels are $k$ and $\mathcal{X}_{t,k}$ is the subset of $\mathcal{X}_{t}$ including all the target samples annotated as class $k$ by the teacher classifier $\tilde h$. This loss is slightly different from the original feature matching loss: it matches the statistics of the representation space $\mathcal{F}$ which totally 
determine the predictions instead of those produced by an extra critic network. Arguably, the objective has a local optima where class-conditional distributions are matched thoroughly. The cluster alignment loss and the discriminative clustering loss work together to align the class-conditional structures of the two domains in a discriminative way.



\subsubsection{Improved marginal distribution alignment}

In fact, the source domain and target domain in the existing popular UDA tasks (\eg, digits adaptation and \emph{Office-31}) have analogous marginal distributions. Therefore, in these experiments, we combine CAT with the marginal distribution alignment methods
, and CAT contributes to bias them to match the cluster-based marginal distributions. The negative effects of these methods of ignoring the discriminability may hurt the stability of training and the capability of converged models. For example, several target circle samples in Fig.~\ref{fig:0}-left will be misclassified by 
them. 

We are dedicated to delivering a technique to improve these models given the observation that in the early stages of training, a portion of target samples lie around the decision boundaries of the adapted classifier, \ie, they have low classification confidence (the largest output probability) and are likely to be misclassified. Therefore, these samples are possible to be mapped into the incorrect clusters in the marginal alignment process and the training falls into local optima. To solve this, we propose to use confidence-thresholding method to hold out uncertain data points with confidence less than $p$, while aligning the confident instances which are more geometrically close to the source domain with the source data. 
Formally, we instantiate this technique in the typical and brief RevGrad~\cite{ganin2015unsupervised} and propose robust RevGrad (rRevGrad) which optimizes the following loss:
\begin{multline}
    \min_{\theta}\max_{\phi}  \mathcal{L}_{d}(\mathcal{X}_s, \mathcal{X}_t) =  
    \frac{1}{N}\sum_{i=1}^N\left[ \log c\left(f(x_s^i;\theta);\phi\right)\right] + \\
    \frac{1}{\tilde M}\sum_{i=1}^{\tilde M}\left[ \log\left(1-c\left(f(x_t^i;\theta);\phi\right)\right)  \gamma_i\right],
\end{multline}
where $c$ is the critic model parameterized by $\phi$ and $\gamma_i$ is an indicator function which outputs 1 only if teacher's confidence of $x_t^i$ is greater than $p$. 
With the divergence between the two domains decreasing,
more and more target samples are selected into the domain adversarial training. Gradually, almost all the target samples are included in the training which avoids the lost of target information. We empirically observed that rRevGrad improves the classification performance on the target domain and enhances the stability of training (see Sec.\ref{sec.exp}). 

\subsection{Discussion}
\textbf{Comparison with SSL based deep UDA methods~\cite{shu2018dirt,saito2017asymmetric,french2017self}.}
CAT not only implements of \emph{cluster} assumption for better classification but also imposes a class-conditional alignment between domains which is more principal in UDA. However, \cite{shu2018dirt,saito2017asymmetric,french2017self} focus on improving classifier to make it more consistent and robust for the target domain based on \emph{cluster} assumption. Thus CAT is compatible to these methods (see Sec.~\ref{sec.exp}). 

\textbf{Comparison with MSTN~\cite{xie2018learning}.}
The cluster alignment loss using conditional feature matching technique is similar to the semantic loss in MSTN~\cite{xie2018learning}. However, in MSTN, minimizing distance between centers is necessary but not sufficient to achieve semantic alignment. In CAT, we regularize the features to form separable and tight clusters, so the feature matching based loss can match the clusters naturally. They are also different in implementation.


\textbf{Mini-batch stochastic training of CAT.}
We implement the two objectives in CAT using stochastically sampled mini-batches as $\mathcal{X}_s$ and $\mathcal{X}_t$. Specifically, $\mathcal{L}_c$ is an instance-wise loss and can work well. The class-conditional expectation $\lambda_{s,k}$ or $\lambda_{t,k}$ in $\mathcal{L}_c$ could be none when these is no points belonging to class $k$. At this time, we remove the term corresponding to class $k$ in Eq.~\ref{eq:(6)} and calculate the mean of the other terms. We empirically find CAT needs only 
$0.05\times$ more training time on a GTX 1080Ti when combining with existing methods.

\textbf{Teacher-student paradigm.}
First, using teacher as labeling function on target domain avoids the error amplification issue. Furthermore, once the classifier becomes more accurate on the target domain, the teacher classifier performs better as well. Then the pseudo labels
used in $\mathcal{L}_c$ and $\mathcal{L}_a$ are more likely to be correct which in turn enhances the classifier. Consequently, a boosting cycle between them is formed.

\section{Experiments}
\label{sec.exp}
To demonstrate the effectiveness of CAT, we evaluate it through various experiments on synthetic imbalanced dataset and three challenging UDA tasks: \emph{SVHN-MNIST-USPS}, \emph{Office-31}~\cite{saenko2010adapting} and \emph{ImageCLEF-DA}\footnote{Source code is at \url{https://github.com/thudzj/CAT}.}. We show that CAT considers and exploits the fine-level class-conditional structures of the source and target domains, and makes the learned feature space discriminative and aligned
, thus yielding improved performance on the target domain.

\textbf{Datasets and configurations.}  
\emph{SVHN-MNIST-USPS} is a challenging adaptation task of digits between three datasets \emph{SVHN}~\cite{netzer2011reading}, \emph{MNIST}~\cite{lecun1998gradient} and \emph{USPS}. We conduct experiments in three directions: \emph{SVHN}$\rightarrow$\emph{MNIST}, \emph{MNIST}$\rightarrow$\emph{USPS} and \emph{USPS}$\rightarrow$\emph{MNIST}. We follow the protocol in~\cite{tzeng2017adversarial}: we use the whole training sets for the adaptation from \emph{SVHN} to \emph{MNIST} and randomly sample 2000 images from \emph{MNIST} and 1800 images from \emph{USPS} for the adaptation between the two datasets. Following MSTN~\cite{xie2018learning}, the images are cast to $28\times28\times1$ when using LeNet~\cite{lecun1998gradient} as classifier. When combining with MCD~\cite{Saito_2018_CVPR} and VADA~\cite{shu2018dirt}, We take the identical settings as the original methods.

\emph{Imbalanced SVHN-MNIST-USPS}. We randomly sample 1000 instances from class 0 and 100 instances from class 1 from the original source domain and construct a new one. Then we sample 100 instances from class 0 and 1000 instances from class 1 from the target domain to form a new target. Therefore, the synthetic adaptation dataset contains several imbalanced two-class adaptation tasks. The experiment settings are the same as those of \emph{SVHN-MNIST-USPS}.

\emph{Office-31} and \emph{ImageCLEF-DA} are two real-world datasets which are widely used in domain adaptation research. \emph{Office-31} is composed of three domains: Amazon (A), DSLR (D) and Webcam (W), containing 2817, 498 and 795 images from 31 categories, respectively. \emph{ImageCLEF-DA} includes three domains: Caltech-256 (C), ImageNet ILSVRC 2012 (I) and Pascal VOC 2012 (P), containing 600 images from 12 classes, respectively. We use data augmentation such as random flipping and cropping in training for fair comparison with the baselines.

\textbf{Implementation.} In synthetic and digits experiments using LeNet, we set $m=30$ according to the performance of CAT on the synthetic dataset and we forward-propagate a target sample twice under different perturbations(\ie, dropout) and use the latter as the prediction of the teacher for its simplicity (similar with the $\Pi$ model of~\cite{laine2016temporal}). In all the other experiments, we fix $m=3$ and deploy a temporal ensemble~\cite{laine2016temporal} of previous predictions of $h$ as teacher (the accumulation decay constant is set to 0.6). We design a ramp-up function similar with that of~\cite{laine2016temporal} to update $\alpha$ in experiments using LeNet and set $\alpha=\frac{2}{1+exp(-10t)}-1$ suggested by RevGrad~\cite{ganin2015unsupervised} in which $t$ increases linearly from 0 to 1 in the others. We set $p=0.9$ in all the experiments without tuning. Refer to Appendix. E for more details of the used architectures and optimization settings.

\subsection{Experiments on \emph{imbalanced SVHN-MNIST-USPS}}
\label{sec-exp-toy}
We first test CAT on the \emph{imbalanced SVHN-MNIST-USPS} dataset, a challenging task where the source domains have $10:1$ ratio of class imbalance while the target domains have $1:10$. We implement CAT and RevGrad~\cite{ganin2015unsupervised} based on the official codes 
of MSTN~\cite{xie2018learning} using LeNet~\cite{lecun1998gradient}. The results are shown in Table~\ref{table:unbalance}. We repeat each task 3 times and report the averaged test accuracy and standard deviation.

\begin{table}[t]

  \centering
  \setlength\tabcolsep{2.5pt}
  \begin{tabular*}{0.48\textwidth}{p{2.cm}ccc}
    \toprule
    \multirow{ 1}{*}{\textbf{Method}}     & \emph{\footnotesize{SVHN} to \footnotesize{MNIST}}  & \emph{\footnotesize{MNIST} to \footnotesize{USPS}}  & \emph{\footnotesize{USPS} to \footnotesize{MNIST}}\\
    \midrule
    \textbf{RevGrad}~\cite{ganin2015unsupervised} & $27.4 \pm 6.3$ &$26.7\pm2.0$ & $17.9\pm1.4$\\
    \textbf{MSTN}~\cite{xie2018learning} & $25.8\pm3.6$ &$30.3\pm1.0$ & $29.4\pm0.5$\\
    \textbf{CAT} & $\mathbf{100.0}\pm0.05$ &$\mathbf{100.0}\pm0.0$ & $\mathbf{99.9}\pm0.2$\\

    \bottomrule
  \end{tabular*}\vspace{-0.2cm}
  \caption{Summary of domain adaptation results on the imbalanced digits datasets in terms of test accuracy (\%).}
   \vspace{-0.3cm}
  \label{table:unbalance}
\end{table}
\begin{table*}[t]
  \vspace{-0.2cm}
  \centering
  \begin{tabular*}{\textwidth}{p{3.5cm}ccccccc}
    \toprule
    \textbf{Method} & A to W & D to W &W to D&A to D&D to A&W to A& Avg \\
    \midrule
    \textbf{ResNet-50}~\cite{he2016deep} & $68.4\pm0.2$ & $96.7\pm0.1$ &$99.3\pm0.1$&$68.9\pm0.2$&$62.5\pm0.3$&$60.7\pm0.3$& $76.1$\\
    \textbf{DAN}~\cite{long2015learning}&$80.5\pm0.4$&$97.1 \pm 0.2$&$ 99.6 \pm 0.1$&$ 78.6\pm0.2$&$63.6\pm 0.3$&$ 62.8 \pm0.2$&80.4\\
    \textbf{RevGrad}~\cite{ganin2015unsupervised} & $82.0\pm0.4$ & $96.9\pm0.2$ &$ 99.1\pm0.1$&$79.4\pm0.4$&$68.2\pm0.4$&$67.4\pm0.5$& $82.2$\\
    \textbf{JAN}~\cite{long2016deep} & $85.4\pm0.3$ & $ 97.4\pm0.2$
    &$99.8\pm0.2$&$84.7\pm0.3$&$68.6\pm0.3$&$70.0\pm0.4$& $84.3$\\
    \textbf{SimNet}~\cite{Pinheiro_2018_CVPR}&$88.6 \pm 0.5$ & $98.2 \pm 0.2$ & $99.7 \pm 0.2$ &$85.3 \pm 0.3$ &$\mathbf{73.4} \pm 0.8$&$\mathbf{71.8} \pm 0.6$&$86.2$\\
    \textbf{GenToAdapt}~\cite{Sankaranarayanan_2018_CVPR}&$89.5\pm 0.5$&$97.9\pm 0.3$&$99.8\pm0.4$&$87.7\pm 0.5$&${72.8}\pm0.3$&${71.4}\pm 0.4$&$86.5$\\
    \textbf{CAT}
    & $91.1\pm0.2$ & $\mathbf{98.6}\pm0.6$ &$99.6\pm0.1$&$90.6\pm1.0$&$70.4\pm0.7$&$66.5\pm0.4$& $86.1$\\
    \textbf{JAN+CAT} & ${94.0} \pm 0.4$ & ${96.6}\pm0.6$ &$\mathbf{100.0}\pm0.0$&${88.1}\pm1.0$&${68.9}\pm0.7$&${69.4}\pm0.5$& ${86.2}$\\
    \textbf{rRevGrad+CAT} & $\mathbf{94.4} \pm 0.1$ & ${98.0}\pm0.2$ &$\mathbf{100.0}\pm0.0$&$\mathbf{90.8}\pm1.8$&${72.2}\pm0.6$&${70.2}\pm0.1$& $\mathbf{87.6}$\\
    \toprule
    \textbf{AlexNet}~\cite{krizhevsky2012imagenet} & $61.6\pm0.5$ & $95.4\pm0.3$ &$99.0\pm0.2$&$63.8\pm0.5$&$51.1\pm0.6$&$49.8\pm0.4$& $70.1$\\
    \textbf{DDC}~\cite{tzeng2014deep} & $ 61.8\pm0.4$ & $ 95.0\pm0.5$ &$ 98.5\pm0.4$&$ 64.4\pm0.3$&$ 52.1\pm0.6$&$52.2\pm0.4$& $ 70.6$\\
    \textbf{DRCN}~\cite{ghifary2016deep} & $68.7\pm0.3$ & $ 96.4\pm0.3$ &$99.0\pm0.2$&$66.8\pm0.5$&$56.0\pm0.5$&$54.9\pm0.5$& $73.6$\\
    \textbf{RevGrad}~\cite{ganin2015unsupervised} & $73.0\pm0.5$ & $96.4\pm0.3$ &$ 99.2\pm0.3$&$72.3\pm0.3$&$53.4\pm0.4$&$51.2\pm0.5$& $74.3$\\
    \textbf{JAN}~\cite{long2016deep} & $74.9\pm0.3$ & $ 96.6\pm0.2$ &$99.5\pm0.2$&$71.8\pm0.2$&$58.3\pm0.3$&$55.0\pm0.4$& $76.0$\\
    \textbf{MSTN}~\cite{xie2018learning} & $80.5\pm0.4 $ & $96.9\pm0.1$ &$99.9\pm0.1$&$74.5\pm0.4$&$62.5\pm0.4$&$60.0\pm0.6$& $79.1$\\
    \textbf{CAT}&$77.4\pm0.1$&$97.4 \pm 0.1$&$ 99.9 \pm 0.1$&$ 74.7\pm0.1$&$63.4\pm 0.2$&$ 60.8 \pm0.6$&78.9\\
    \textbf{JAN+CAT} & $78.4\pm0.5$&$97.2 \pm 0.2$&$ \mathbf{100.0} \pm 0.0$&$ 74.5\pm0.5$&$63.6\pm 0.4$&$ 61.2\pm 0.6$ & $79.2$\\
    \textbf{rRevGrad+CAT} & $\mathbf{80.7} \pm 1.6$ & $\mathbf{97.6}\pm0.1$ &$\mathbf{100.0}\pm0.0$&$\mathbf{76.4}\pm0.6$&$\mathbf{63.7}\pm0.5$&$\mathbf{62.2}\pm0.4$& $\mathbf{80.1}$\\
    \bottomrule
  \end{tabular*}\vspace{-0.2cm}
  \caption{Accuracy on the \emph{Office-31} datasets in terms of test accuracy (\%) (ResNet-50 and AlexNet).}\vspace{-0.2cm}
  \label{table:2}
\end{table*}

\begin{table}[t]

  \centering
  \setlength\tabcolsep{1.5pt}
  \begin{tabular*}{0.48\textwidth}{p{2.4cm}ccc}
    \toprule
    \multirow{ 1}{*}{\textbf{Method}}     & \emph{\footnotesize{SVHN} to \footnotesize{MNIST}}  & \emph{\footnotesize{MNIST} to \footnotesize{USPS}}  & \emph{\footnotesize{USPS} to \footnotesize{MNIST}}\\
    \midrule
    \textbf{Source Only} & $60.1\pm1.1$ &$75.2\pm1.6$ & $57.1\pm1.7$\\
    \textbf{DDC}~\cite{tzeng2014deep} & $68.1\pm0.3$ &$79.1\pm0.5$ & $66.5\pm3.3$\\
    \textbf{CoGAN}~\cite{liu2016coupled} & - &$91.2\pm0.8$ & $89.1\pm0.8$\\
    \textbf{DRCN}~\cite{ghifary2016deep} & $82.0\pm0.1$ &$91.8\pm0.09$ & $73.7\pm0.04$\\
    \textbf{ADDA}~\cite{tzeng2017adversarial} & $76.0\pm1.8$ &$89.4\pm0.2$ & $90.1\pm0.8$\\
    \textbf{LEL}~\cite{luo2017label} & $81.0\pm0.3$ &- & -\\
    \textbf{AssocDA}~\cite{haeusser2017associative} & $97.6$ & - & - \\
    \textbf{MSTN}~\cite{xie2018learning} & $91.7\pm1.5$ &$92.9\pm1.1$ & -\\
    \textbf{CAT} & $98.1\pm1.3$ &$90.6\pm2.3$ & $80.9\pm3.1$\\
    \midrule
    \textbf{RevGrad}~\cite{ganin2015unsupervised} & $73.9$ &$77.1\pm1.8$ & $73.0\pm2.0$\\
    \textbf{RevGrad+CAT} & $98.0\pm0.8$ &$93.7\pm1.1$ & $95.7\pm1.3$\\
    \textbf{rRevGrad+CAT} & $\mathbf{98.8}\pm0.02$ &$94.0\pm0.7$ & $\mathbf{96.0}\pm0.9$\\
    
    \midrule
    \textbf{MCD}~\cite{Saito_2018_CVPR} & $96.2\pm0.4$ &$94.2\pm0.7$ & $94.1\pm0.3$\\
    \textbf{MCD+CAT} & $97.1\pm0.2$ &$\mathbf{96.3}\pm0.5$ & $95.2\pm0.4$\\
    \midrule
    \textbf{VADA}~\cite{shu2018dirt} & $94.5$ & - & -\\
    \textbf{VADA+CAT} & $95.2$ & - & -\\
    \bottomrule
    
  \end{tabular*}\vspace{-0.2cm}
  \caption{Summary of domain adaptation results on the digits datasets in terms of test accuracy (\%).}
   \vspace{-0.4cm}
  \label{table:1}
\end{table}
It is notable that RevGrad~\cite{ganin2015unsupervised} and MSTN~\cite{xie2018learning} fail thoroughly owing to their obsession of matching the marginal distributions. In contrast, CAT gives almost completely correct predictions for the target domains. 
This experiment verifies that existing methods through aligning marginal distributions are restrictive and require the modes corresponding to the same class but different domains to be geometrically similar. They are sensitive and fragile in practical tasks.

\subsection{\emph{SVHN-MNIST-USPS} digits datasets}

We apply CAT to the popular digits adaptation task \emph{SVHN-MNIST-USPS} and compare to the state-of-the-art approaches in Table~\ref{table:1} (all baseline results are taken from related literature). CAT, RevGrad+CAT and rRevGrad+CAT follow the settings of MSTN~\cite{xie2018learning} using the LeNet~\cite{lecun1998gradient}. We implement MCD+CAT and VADA+CAT based on the official codes of MCD~\cite{Saito_2018_CVPR} 
and VADA~\cite{shu2018dirt} 
using their architectures instead of LeNet for fair comparison. We only integrate CAT with the first stage algorithm VADA in DIRT-T~\cite{shu2018dirt} while discarding the fine-tuning stage for simpleness.

There are several conclusions we can make. First, CAT reveals strikingly improved test accuracy on \emph{SVHN} to \emph{MNIST} task without tuning the hyper-parameters, and CAT even outperforms MCD~\cite{Saito_2018_CVPR} and VADA~\cite{shu2018dirt} which use much wider and deeper neural networks thanks to the class-conditional discriminative alignment between the source and target domains. This task is the most challenging one among the three because of the complex samples and the internal class imbalance of \emph{SVHN}. Second, CAT does not perform well enough on the other two tasks but when combining with rRevGrad and MCD~\cite{Saito_2018_CVPR}, CAT outperforms the strong baselines MSTN~\cite{xie2018learning} and MCD~\cite{Saito_2018_CVPR} with obvious margins. Third, applying CAT into RevGrad~\cite{ganin2015unsupervised}, MCD~\cite{Saito_2018_CVPR} and VADA~\cite{shu2018dirt} can enhance the base methods significantly, especially on the typical and simple RevGrad~\cite{ganin2015unsupervised}. Finally, rRevGrad+CAT displays higher test accuracy and lower variance than those of RevGrad+CAT and the advantage is particularly obvious when the two domains have different class-conditional structures (\eg, \emph{SVHN} to \emph{MNIST}), so we utilize rRevGrad+CAT on more challenging tasks.

\vspace{-0.cm}
\subsection{Experiments on \emph{Office-31} and \emph{ImageCLEF-DA}}
\vspace{-0.1cm}

We evaluate CAT using two sets of extensive experiments on the widely used \emph{Office-31} and \emph{ImageCLEF-DA}. They contain more realistic and high-dimensional images, providing a good complement to the digits adaptation task. The results are provided in Table~\ref{table:2} and Table~\ref{table:3}, respectively. We integrate CAT with rRevGrad and JAN~\cite{long2016deep} (using ResNet-50~\cite{he2016deep} and AlexNet~\cite{krizhevsky2012imagenet} as the classifiers), which is sufficient to testify the effectiveness of discriminative cluster-based alignment and teacher-student paradigm.

We observe that CAT can boost rRevGrad and JAN~\cite{long2016deep} significantly, especially on the difficult A to W, A to D and D to A tasks in \emph{Office-31}, and the combined models surpass the strong baselines RevGrad~\cite{ganin2015unsupervised} and JAN~\cite{long2016deep} by obvious margins. CAT based methods also outperform MSTN~\cite{xie2018learning} on various tasks substantially which proves the class-conditional discriminative alignment is superior to the semantic alignment used by MSTN~\cite{xie2018learning}. The improvement of test accuracy on most tasks of \emph{ImageCLEF-DA} shows that CAT can still work well when the domains are small containing only 600 images. 
We further confirm that CAT can deliver a discriminative and aligned feature space by visualizing the learned features in the Appendix. A.

\begin{table*}[t]
  \vspace{-0.2cm}
  \centering
  \begin{tabular*}{\textwidth}{p{3.5cm}ccccccc}
    \toprule
    \textbf{Method} & I to P & P to I &I to C&C to I&C to P&P to C& Avg \\
    \midrule
    \textbf{ResNet-50}~\cite{he2016deep} & $74.8 \pm0.3$ & $83.9\pm0.1$ &91.5$\pm0.3$&$78.0\pm0.2$&$ 65.5\pm0.3$&$91.2\pm0.3$& $80.7$\\
     \textbf{DAN}~\cite{long2015learning}&$74.5\pm0.4$&$82.2 \pm 0.2$&$ 92.8 \pm 0.2$&$ 86.3\pm0.4$&$69.2\pm 0.4$&$ 89.8 \pm0.4$&82.5\\
    \textbf{RevGrad}~\cite{ganin2015unsupervised} & $ 75.0\pm0.6 $ & $86.0\pm0.3$ &$\mathbf{96.2}\pm0.4$&$87.0\pm0.5$&$74.3\pm0.5$&$91.5\pm0.6$& $85.0$\\
     \textbf{JAN}~\cite{long2016deep} & $ 76.8\pm0.4$ & $ 88.0\pm0.2$ &$94.7\pm0.2$&$89.5\pm0.3$&$ 74.2\pm0.3$&$91.7\pm0.3$& $85.8$\\
    \textbf{CAT} & $76.7\pm0.2$ & $89.0\pm0.7$ & $94.5\pm0.4$ & $89.8\pm0.3$ & $74.0\pm0.2$ & $\mathbf{93.7}\pm1.0$ & $86.3$ \\
    \textbf{JAN+CAT} & ${76.3}\pm0.8$ & ${89.2}\pm0.8$ &${95.3}\pm0.7$&${89.3}\pm0.3$&$\mathbf{75.9}\pm1.1$&${92.2}\pm1.3$& ${86.4}$\\
    \textbf{rRevGrad+CAT} & $\mathbf{77.2}\pm0.2$ & $\mathbf{91.0}\pm0.3$ &${95.5}\pm0.3$&$\mathbf{91.3}\pm0.3$&${75.3}\pm0.6$&$93.6\pm0.5$& $\mathbf{87.3}$\\
    \toprule
    \textbf{AlexNet}~\cite{krizhevsky2012imagenet} & $ 66.2\pm0.2$ & $70.0\pm0.2$ &$84.3\pm0.2$&$71.3\pm0.4$&$ 59.3\pm0.5$&$84.5\pm0.3$& $73.9$\\
    \textbf{RTN}~\cite{long2016unsupervised} & $ 67.4\pm0.3$ & $81.3\pm0.3$ &$89.5\pm0.4$&$78.0\pm0.2$&$62.0\pm0.2$&$ 89.1\pm0.1$& $77.9$\\
    \textbf{RevGrad}~\cite{ganin2015unsupervised} & $66.5\pm0.5$ & $81.8\pm0.4$ &$ 89.0\pm0.5$&$79.8\pm0.5$&$63.5\pm0.4$&$ 88.7\pm0.4$& $78.2$\\
    \textbf{JAN}~\cite{long2016deep} & $ 67.2\pm0.5$ & $ 82.8\pm0.4$ &$91.3\pm0.5$&$80.0\pm0.5$&$ 63.5\pm0.4$&$91.0\pm0.4$& $79.3$\\
    \textbf{MSTN}~\cite{xie2018learning} & $ 67.3\pm0.3 $ & $82.8\pm0.2$ &$91.5\pm0.1$&$\mathbf{81.7}\pm0.3$&$65.3\pm0.2$&$91.2\pm0.2$& $80.0$\\
    \textbf{CAT}&$68.3\pm0.5$&$83.6 \pm 0.7$&$ 91.3 \pm 0.3$&$ 79.1\pm0.5$&$64.0\pm 0.7$&$ 90.9 \pm0.3$&79.5\\
    \textbf{JAN+CAT} & $67.8\pm0.2$&$83.7 \pm 0.4$&$ \mathbf{92.3} \pm 0.6$&$ 80.8\pm0.3$&$\mathbf{65.8}\pm 0.8$&$ 91.1\pm 0.2$ & $80.3$\\
    \textbf{rRevGrad+CAT} & $\mathbf{68.6}\pm0.1$ & $\mathbf{84.6}\pm0.5$ &$91.9\pm0.4$&${80.8}\pm0.3$&$65.6\pm0.6$&$\mathbf{92.5}\pm0.2$& $\mathbf{80.7}$\\
    \bottomrule
  \end{tabular*}\vspace{-0.2cm}
  \caption{Accuracy on the \emph{ImageCLEF-DA} datasets in terms of test accuracy (\%) (ResNet-50 and AlexNet).}\vspace{-0.2cm}
  \label{table:3}
\end{table*}

\begin{figure}[t]
\vspace{-0.cm}
\centering
\begin{subfigure}{0.2\textwidth}
  \centering
  \includegraphics[width=\linewidth]{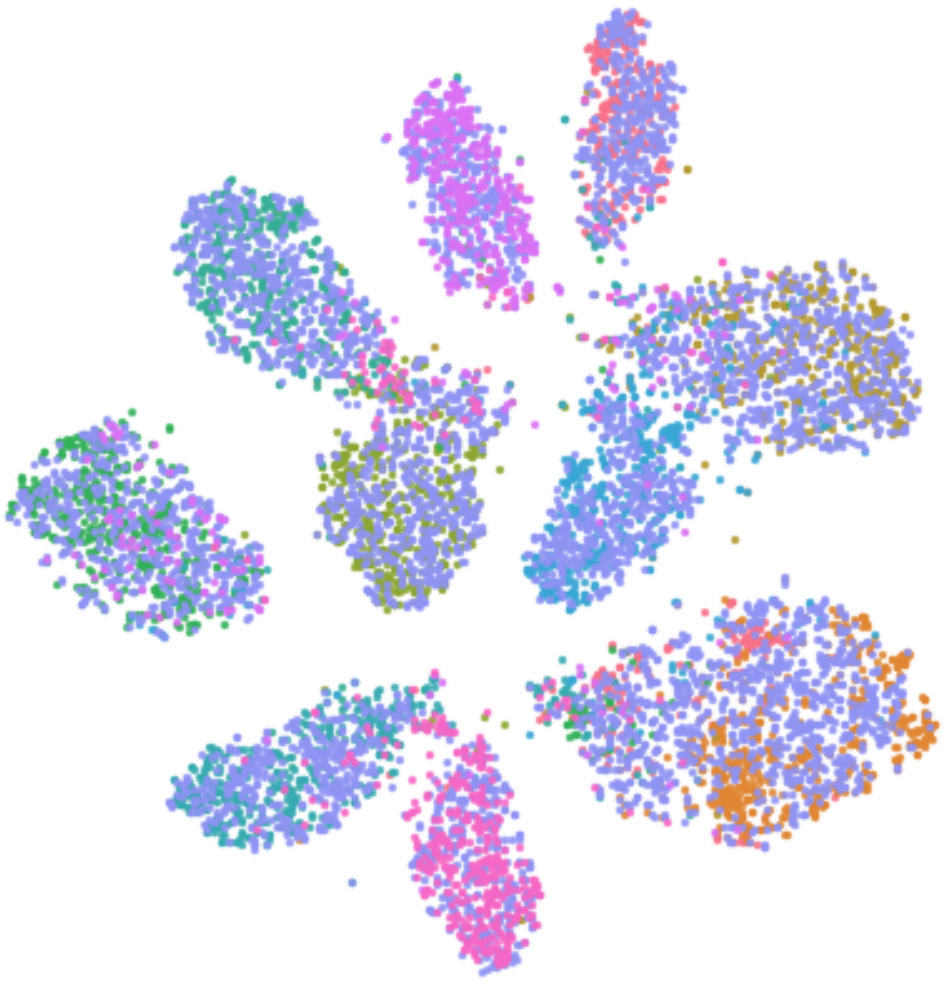}\vspace{-0.1cm}
  \caption{RevGrad}
  \label{fig:2-1}
\end{subfigure}
\begin{subfigure}{0.2\textwidth}
  \centering
  \includegraphics[width=\linewidth]{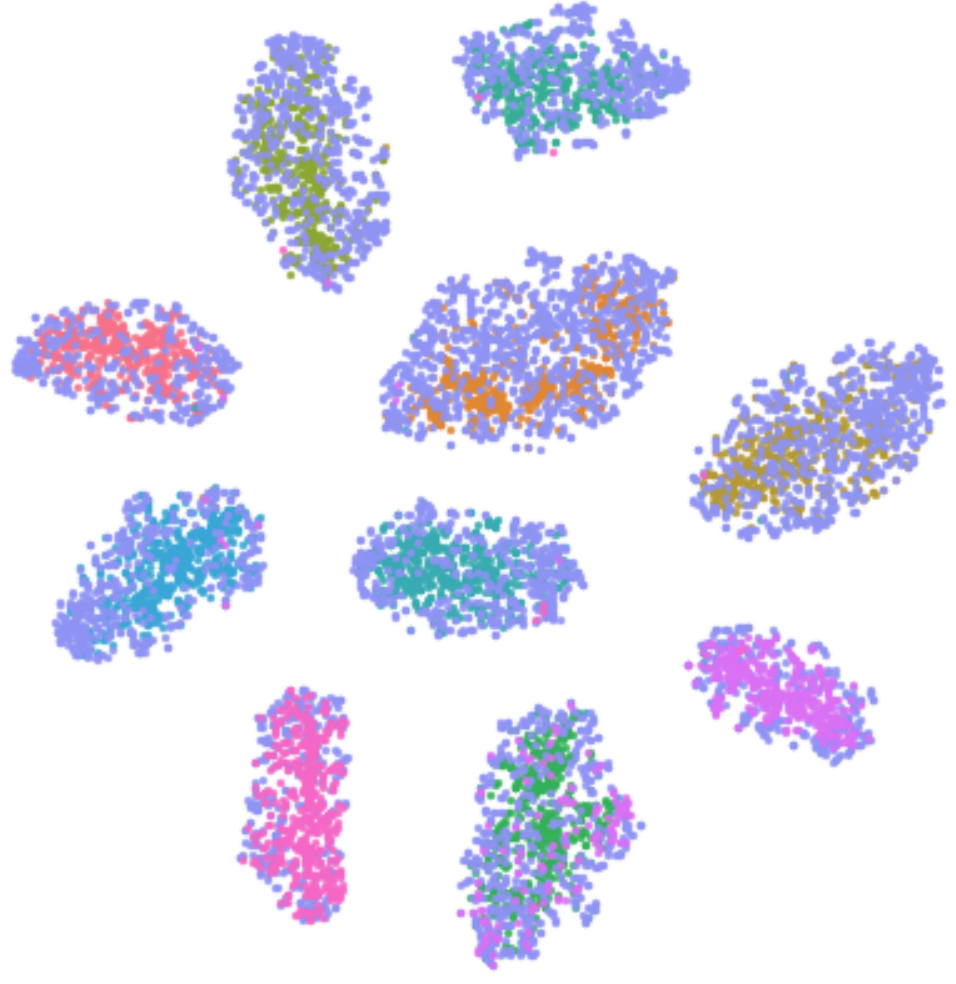}\vspace{-0.1cm}
  \caption{rRevGrad+CAT}
  \label{fig:2-2}
\end{subfigure}
\vspace{-0.2cm}
\caption{(Best viewed in color.) (a) Feature space learned by RevGrad. (b) Feature space learned by rRevGrad+CAT. The features are projected to 2-D using t-SNE. Blue violet denotes the source domain and the other colors denotes classes of target domain. See Appendix. A for more results.}\vspace{-0.cm}
\label{fig:3}
\end{figure}

\vspace{-0.1cm}
\subsection{Analysis}
\vspace{-0.1cm}
\textbf{Visualization of feature space.} 
\label{sec-exp4}
We visualize the features of the two domains learned by the powerful rRevGrad+CAT and RevGrad~\cite{ganin2015unsupervised} on the \emph{SVHN} to \emph{MNIST} 
task using 
t-SNE~\cite{maaten2008visualizing}. The results are shown in Fig.~\ref{fig:3}. As expected, using CAT (Fig.~\ref{fig:2-2}), the features are concentrated and form tight clusters and those from different classes are separated. In contrast, the features learned by RevGrad~\cite{ganin2015unsupervised} (Fig.~\ref{fig:2-1}) are more overlapping and less discriminative.

\begin{figure}[t]
\centering
 \vspace{0.3cm}
  \includegraphics[width=\linewidth]{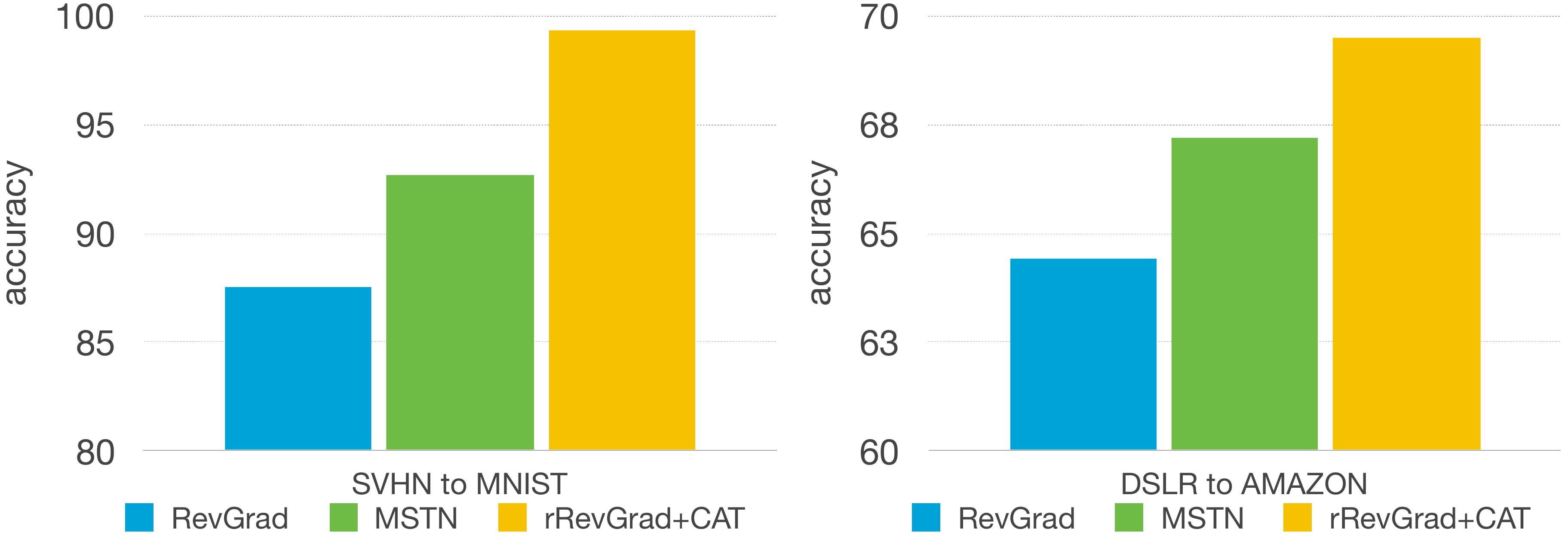}  \vspace{-0.5cm}
\caption{Summary of clustering accuracy(\%).}\vspace{-0.cm}
\label{fig:6}
\end{figure}

\textbf{Clustering in the feature space.}
We further examine the feature space shaped by CAT and other baselines by conducting K-means~\cite{lloyd1982least} clustering using the aligned features. We utilize the trained models to infer the hidden features of all the images from two domains. Then the features are clustered into $k$ components by K-means in scikit-learn~\cite{witten2016data}. We set $k$ as the number of categories. We greedily set the label of a cluster as the most frequent label in it to calculate clustering accuracy as shown in Fig.~\ref{fig:6}. As expected, the feature spaces learned by rRevGrad+CAT demonstrate more discriminative cluster structure and this is consistent with the classification results. Appendix. B, C and D provide more analyses.

\vspace{-0.2cm}
\section{Conclusion}
\vspace{-0.1cm}
In this paper we address the challenges of making better alignment between domains and advocate to exploit the discriminative class-conditional structures for effective adaptation in deep UDA. We propose \emph{Cluster Alignment with a Teacher} (CAT) to achieve the objectives of discriminative learning and class-conditional alignment via a discriminative clustering loss and a cluster-based alignment loss. CAT produces a domain-invariant feature space with improved discriminative power and enhances the performance significantly. CAT establishes new state-of-the-art baselines on benchmarks and additional analyses testify its effectiveness.

 
\vspace{-0.2cm}
\section*{Acknowledgements}
\vspace{-0.2cm}
{\small
This work was supported by the NSFC Projects (Nos. 61621136008, 61620106010),  Beijing NSF Project (No. L172037), Tiangong Institute for Intelligent Computing, Beijing Academy of Artificial Intelligence (BAAI), the NVIDIA NVAIL Program with GPU/DGX Acceleration and the JP Morgan Faculty Research Program.
}

{\small
\bibliographystyle{ieee_fullname}
\bibliography{egbib}
}

\end{document}


\title{Supplementary Material:\\
Cluster Alignment with a Teacher for Unsupervised Domain Adaptation}

\author{Zhijie Deng, Yucen Luo, Jun Zhu\thanks{Corresponding author.}\\
Dept. of Comp. Sci. \& Tech., Institute for AI, BNRist Lab, THBI Lab, Tsinghua University\\
{\tt\small \{dzj17, luoyc15\}@mails.tsinghua.edu.cn, dcszj@tsinghua.edu.cn}
}

\maketitle
\ificcvfinal\thispagestyle{empty}\fi

\appendix

\section{Class-conditional cluster structure on more tasks}

\begin{figure}[t]
\vspace{-0.4cm}
\centering
\begin{subfigure}{0.2\textwidth}
  \centering
  \includegraphics[width=\linewidth]{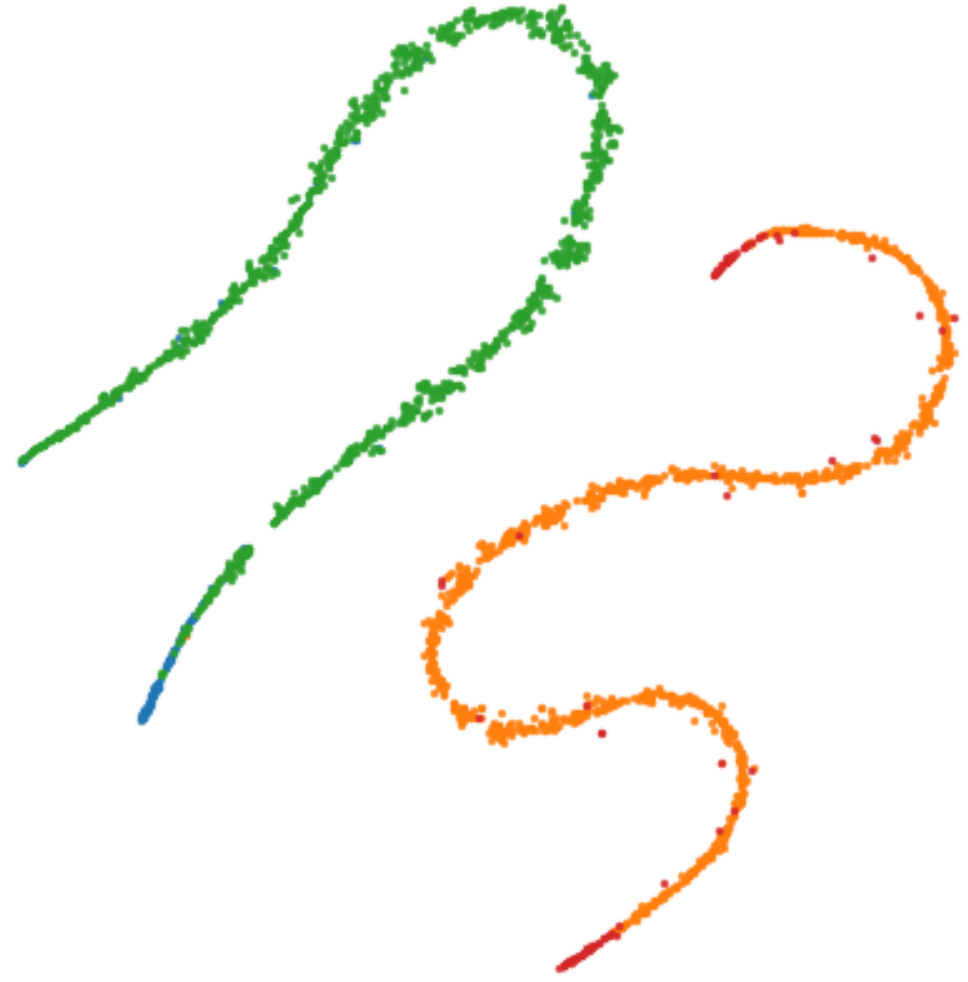}
  \caption{CAT}
  \label{afig:im-3}
\end{subfigure}
\begin{subfigure}{0.2\textwidth}
  \centering
  \includegraphics[width=\linewidth]{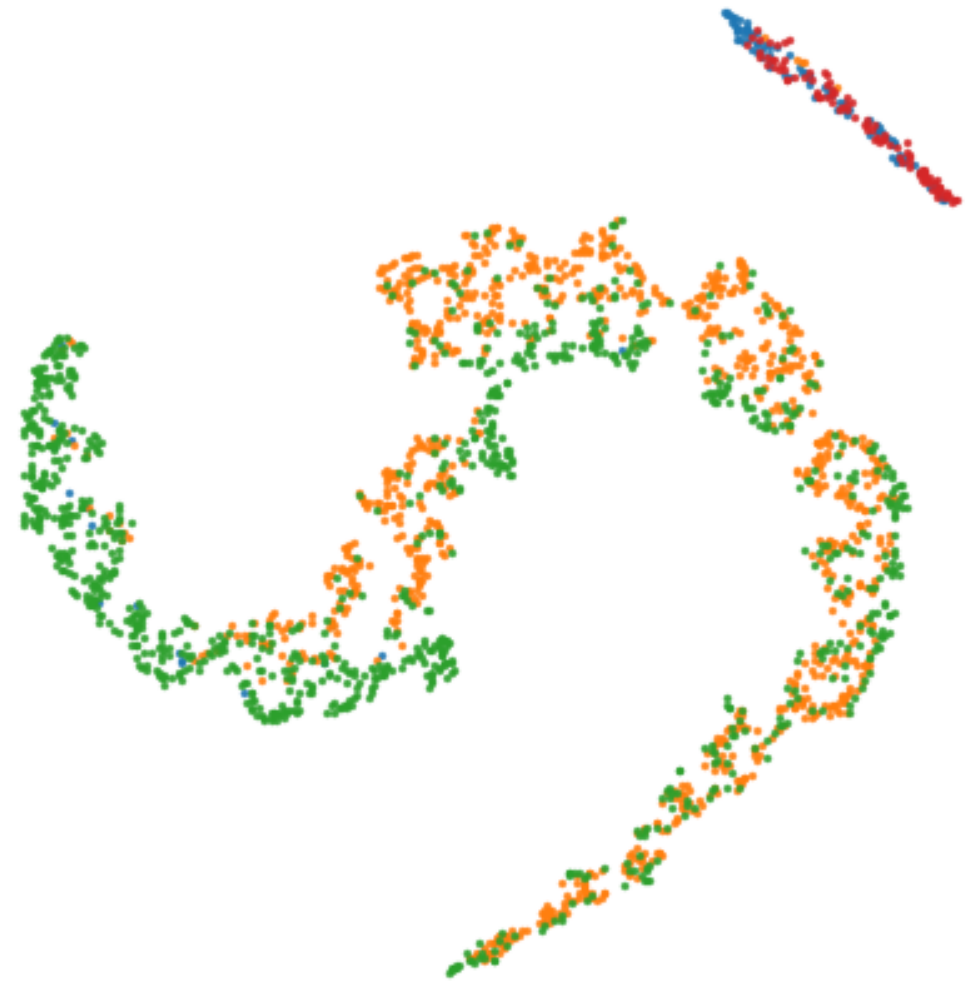}
  \caption{RevGrad}
  \label{afig:im-1}
\end{subfigure}
\begin{subfigure}{0.2\textwidth}
  \centering
  \includegraphics[width=\linewidth]{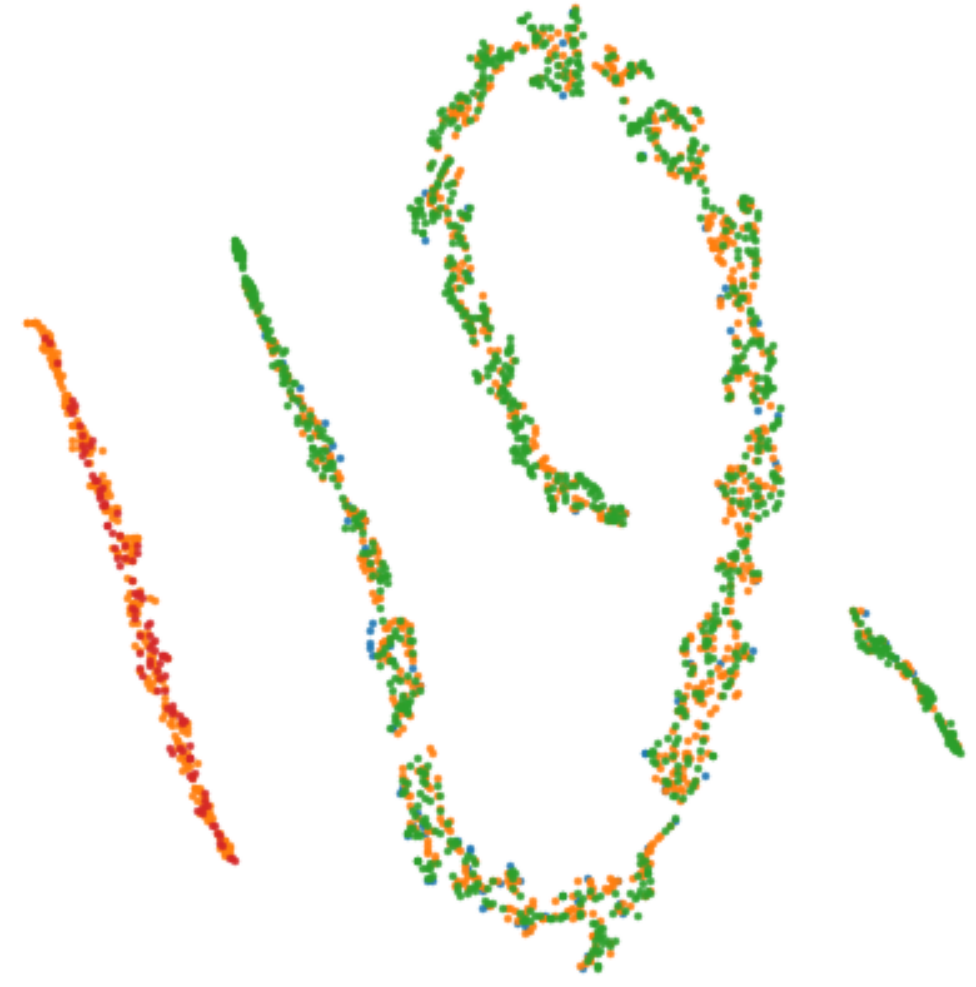}
  \caption{MSTN}
  \label{afig:im-2}
\end{subfigure}
\vspace{-0.2cm}
\caption{(Best viewed in color.) Feature space learned on imbalanced \emph{SVHN} to \emph{MNIST} task. Green, red, blue and orange points represent `0' images from \emph{SVHN}, `1' images from \emph{SVHN}, `0' images from \emph{MNIST} and `1' images from \emph{MNIST}, respectively.}
\vspace{-0.4cm}
\label{afig:1}
\end{figure}

First, we visualize the learned feature spaces of CAT, RevGrad [7] and MSTN [49] on the imbalanced \emph{SVHN} to \emph{MNIST} task using t-SNE [27], as shown in Fig.\ref{afig:1}. It is obvious that CAT can force the samples from the same class to concentrate together to form tighter clusters than those of RevGrad and MSTN, and the clusters present strip pattern in the 2-D space. CAT can also align the class-conditional distributions of the source and the target domains correctly. However, RevGrad and MSTN tend to align the `0' images in \emph{SVHN} with the `1' images in \emph{MNIST}, thus the learned feature spaces of them are confusing and not discriminative. This visualization verifies the results in Table. 1.

\begin{figure}[t]
\centering
\vspace{-0.3cm}
\begin{subfigure}{0.2\textwidth}
  \centering
  \includegraphics[width=\linewidth]{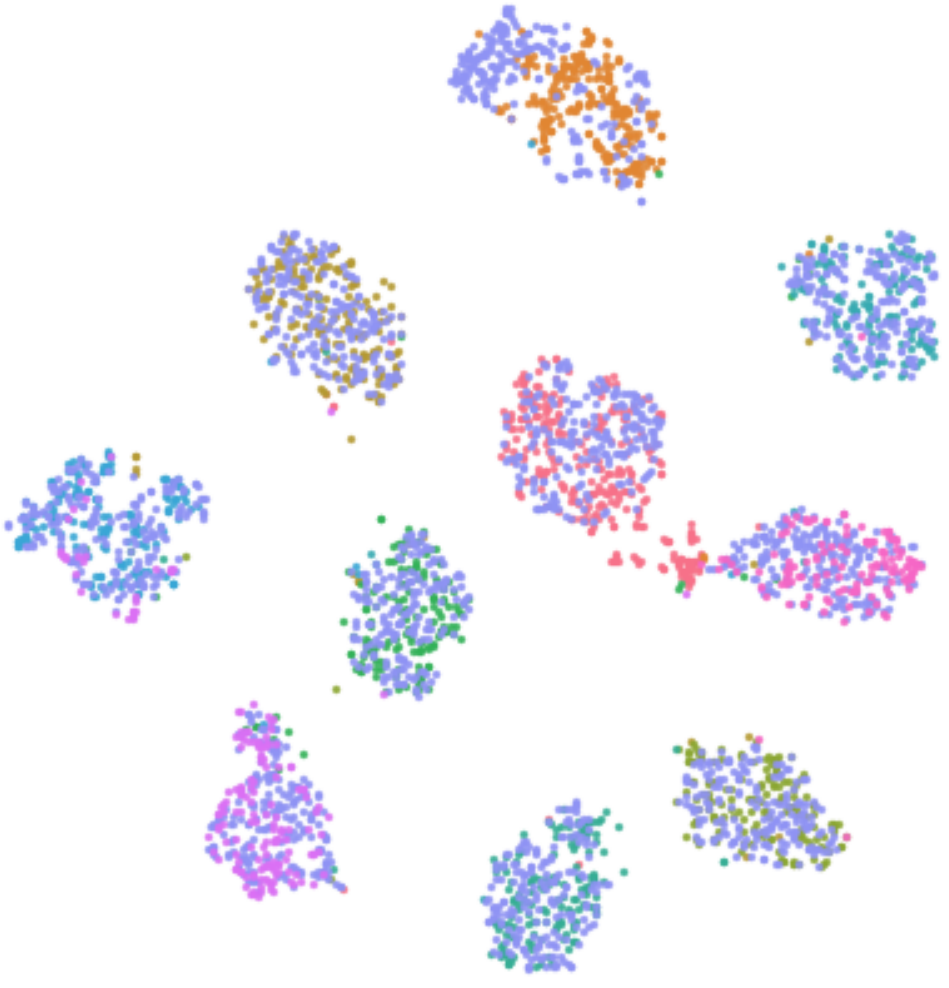}
  \caption{RevGrad}
  \label{afig:2-1}
\end{subfigure}
\begin{subfigure}{0.2\textwidth}
  \centering
  \includegraphics[width=\linewidth]{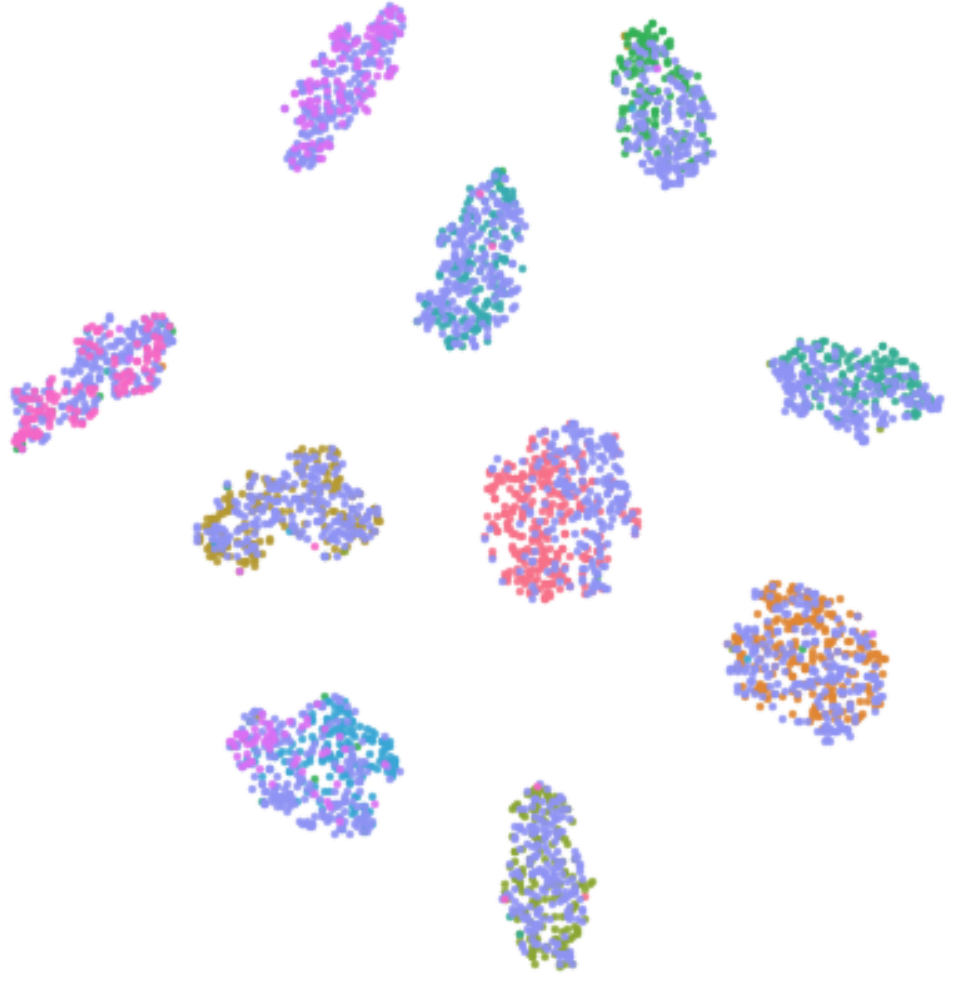}
  \caption{rRevGrad+CAT}
  \label{afig:2-2}
\end{subfigure}
\begin{subfigure}{0.2\textwidth}
  \centering
  \includegraphics[width=\linewidth]{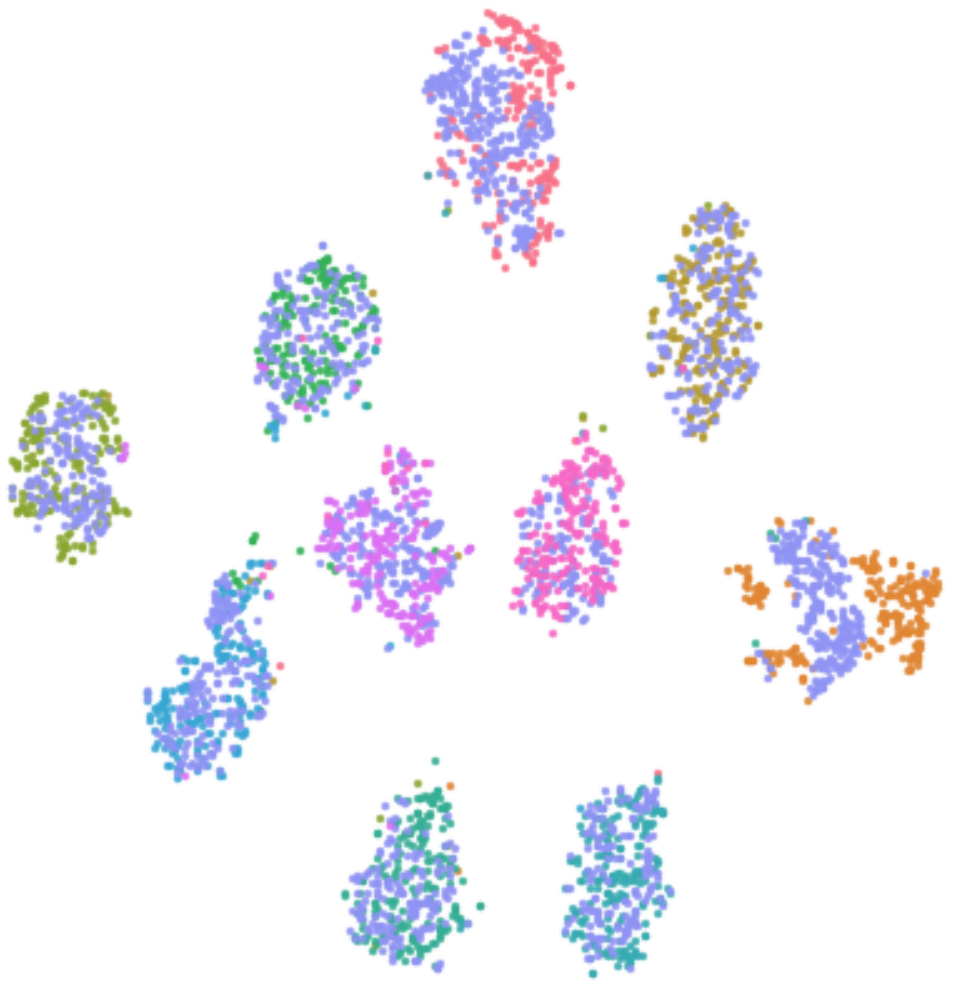}
  \caption{RevGrad}
  \label{afig:3-1}
\end{subfigure}
\begin{subfigure}{0.2\textwidth}
  \centering
  \includegraphics[width=\linewidth]{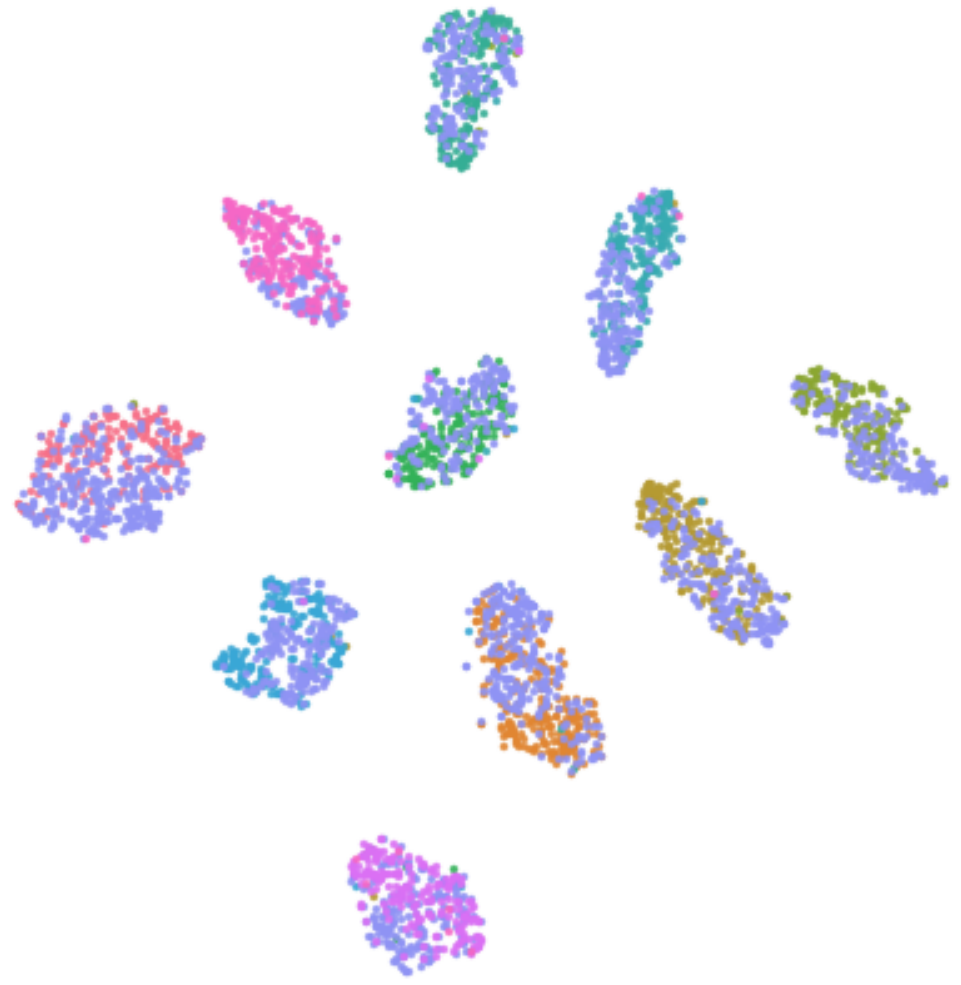}
  \caption{rRevGrad+CAT}
  \label{afig:3-2}
\end{subfigure}
\vspace{-0.2cm}
\caption{(Best viewed in color.) Feature space learned on \emph{MNIST} to \emph{USPS} (Fig.~\ref{afig:2-1} and Fig.~\ref{afig:2-2}) and \emph{USPS} to \emph{MNIST} (Fig.~\ref{afig:3-1} and Fig.~\ref{afig:3-2}) tasks. Blue violet denotes the source domain and the other colors denote different classes of target domain.}
\label{afig:2}
\end{figure}

Second, we plot the feature spaces learned by CAT+rRevGrad and RevGrad on \emph{MNIST} to \emph{USPS} and \emph{USPS} to \emph{MNIST} tasks in Fig.~\ref{afig:2} using t-SNE [27]. CAT+rRevGrad can deliver more discriminative feature spaces with separable and tight class-conditional clusters. Therefore, it is sufficient to use the first-order statistics based matching loss $\mathcal{L}_a$ to match the class-conditional distributions of the two domains. The aligned clusters of the source and the target domains also verify the effectiveness of the loss $\mathcal{L}_a$.

Furthermore, we examine the feature space learned by CAT on more challenging tasks in \emph{Office-31} dataset and \emph{ImageCLEF-DA} dataset, and results are demonstrated in Fig.~\ref{afig:5}. These features are outputs of AlexNet trained with rRevGrad+CAT. The class-conditional distributions are shaped to be tight and separable clusters, and the corresponding cluters from the source domain and the target domain are aligned. Therefore, CAT can achieve the objectives of discriminative learning and class-conditional alignment, thus can perform well on the extensive experiments on \emph{Office-31} and \emph{ImageCLEF-DA} datasets.

\begin{figure}[t]
\vspace{-0.cm}
\centering
\begin{subfigure}{0.2\textwidth}
  \centering
  \includegraphics[width=\linewidth]{a2wcat.pdf}
  \caption{Amazon to Webcam}
  \label{afig:5-1}
\end{subfigure}
\begin{subfigure}{0.2\textwidth}
  \centering
  \includegraphics[width=\linewidth]{a2dcat.pdf}
  \caption{Amazon to DSLR}
  \label{afig:5-2}
\end{subfigure}
\begin{subfigure}{0.2\textwidth}
  \centering
  \includegraphics[width=\linewidth]{p2icat.pdf}
  \caption{p to i}
  \label{afig:5-3}
\end{subfigure}
\begin{subfigure}{0.2\textwidth}
  \centering
  \includegraphics[width=\linewidth]{i2pcat.pdf}
  \caption{i to p}
  \label{afig:5-4}
\end{subfigure}
\vspace{-0.2cm}
\caption{(Best viewed in color.) Feature space learned on four challenging tasks. Blue violet (in (a) and (b)) and deep sky blue (in (c) and (d)) denote the source domain and the other colors denote different classes of target domain.}\vspace{-0.3cm}
\label{afig:5}
\end{figure}

\begin{figure}[t]
\vspace{-0.3cm}
\centering
\begin{subfigure}{0.22\textwidth}
  \centering
  \includegraphics[width=\linewidth]{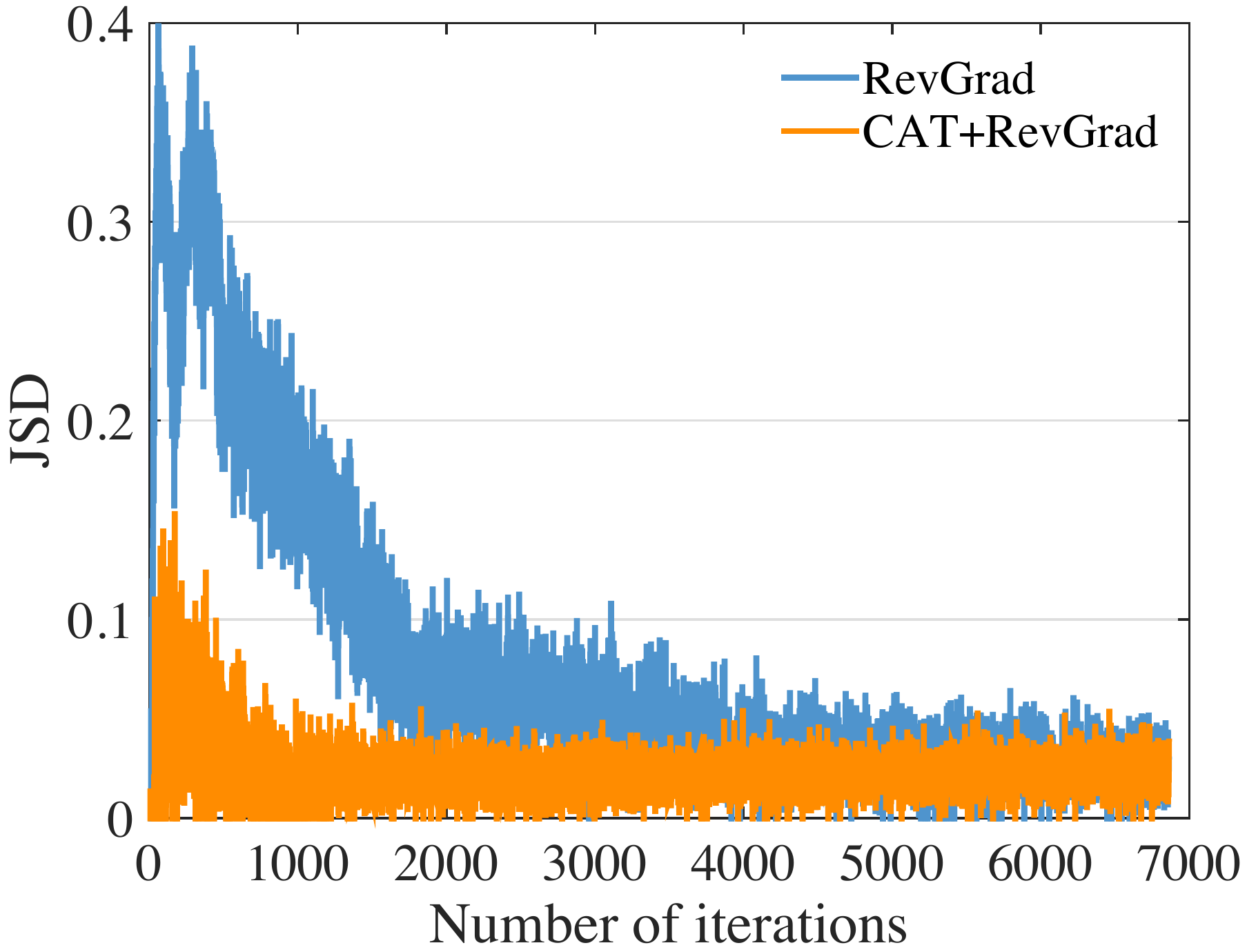}\vspace{-0.1cm}
  \caption{DSLR to Amazon}
  \label{afig:4-1}
\end{subfigure}
\begin{subfigure}{0.22\textwidth}
  \centering
  \includegraphics[width=\linewidth]{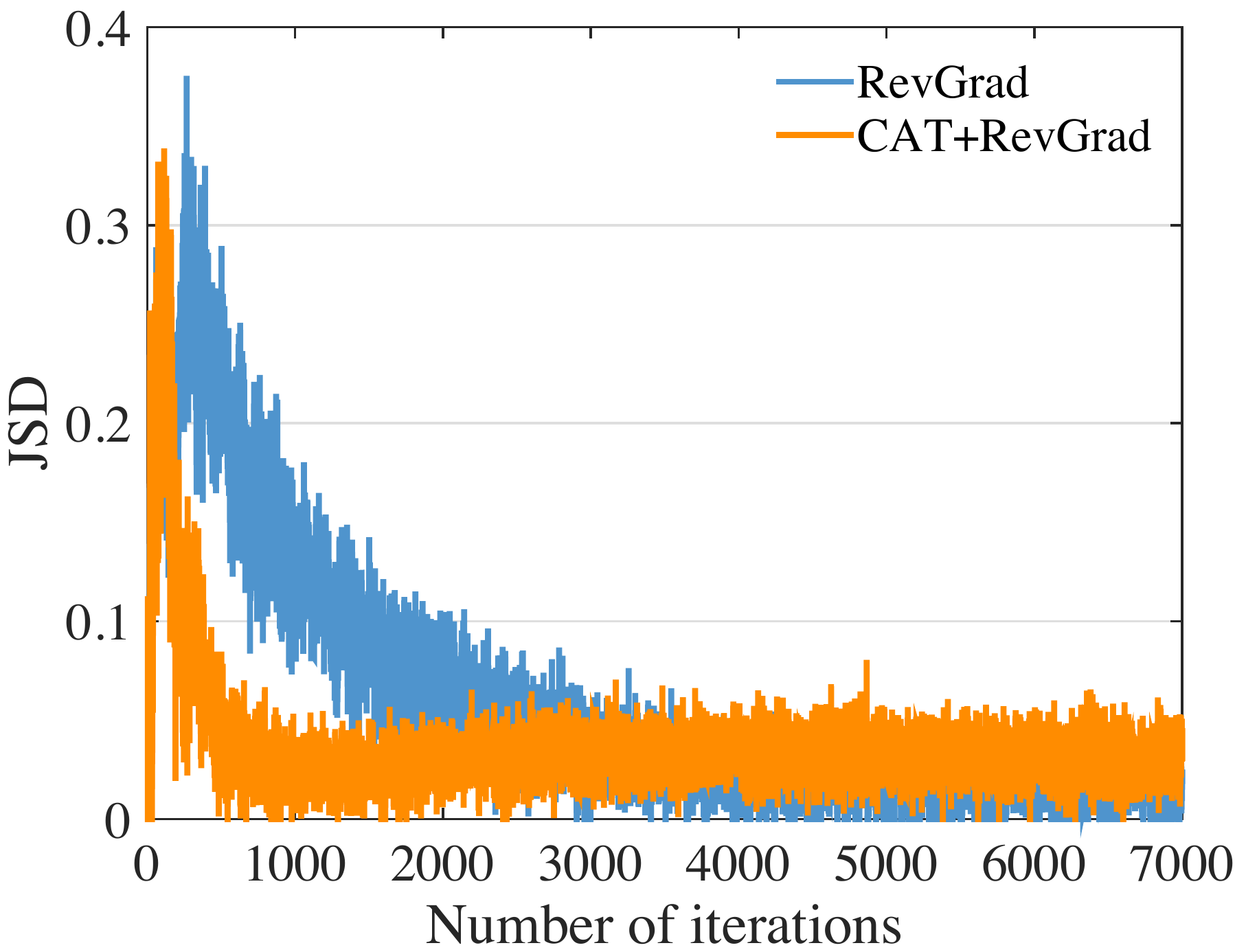}\vspace{-0.1cm}
  \caption{Amazon to DSLR}
  \label{afig:4-2}
\end{subfigure}

\caption{Jensen-Shannon divergence (JSD) curves during training.}
\label{afig:4}
\end{figure}

\begin{figure}[t]

\centering
\begin{subfigure}{0.2\textwidth}
  \centering
  \includegraphics[width=\linewidth]{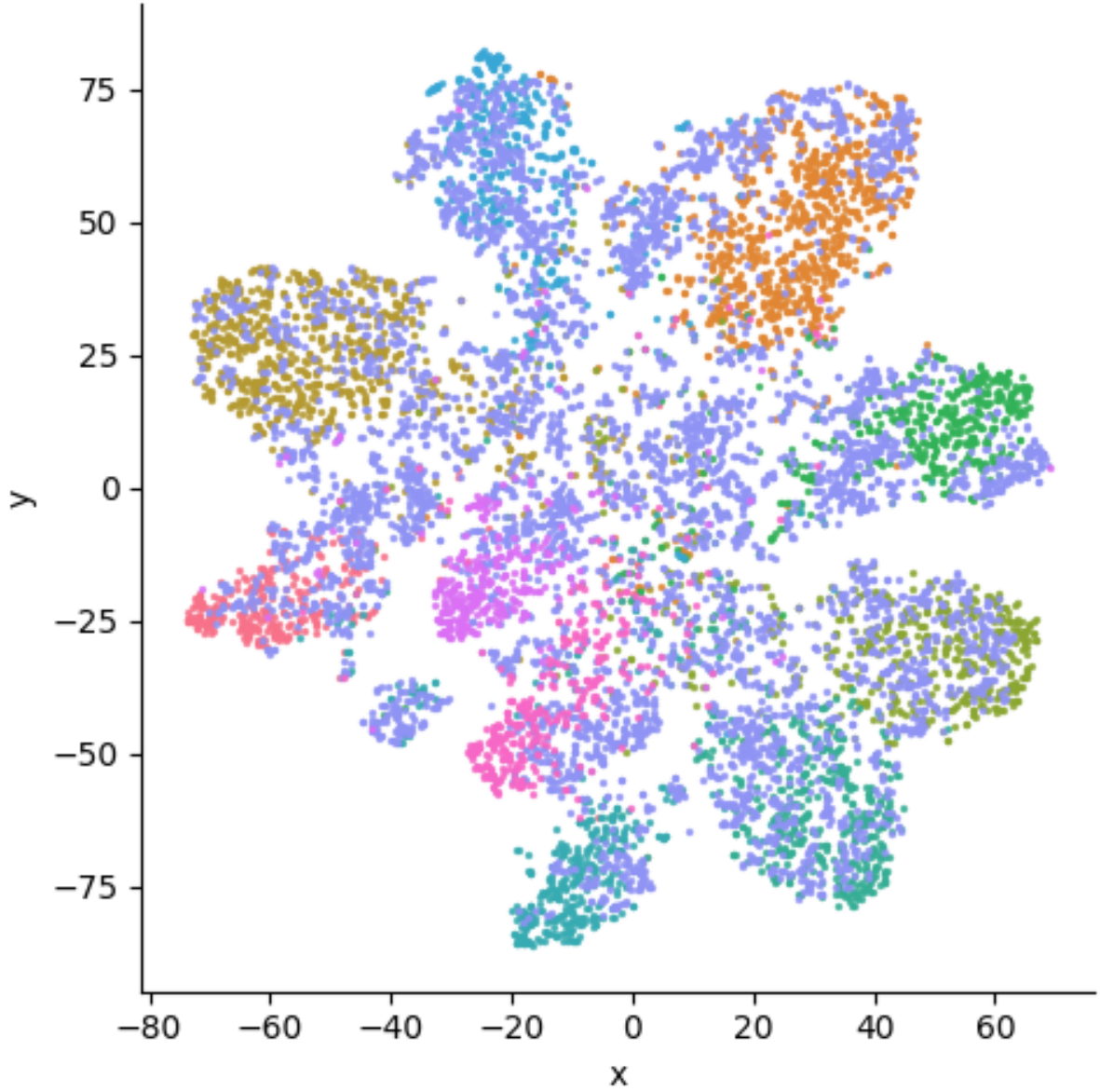}\vspace{-0.1cm}
  \caption{\emph{SVHN} to \emph{MNIST}}
  \label{afig:7-1}
\end{subfigure}
\begin{subfigure}{0.2\textwidth}
  \centering
  \includegraphics[width=\linewidth]{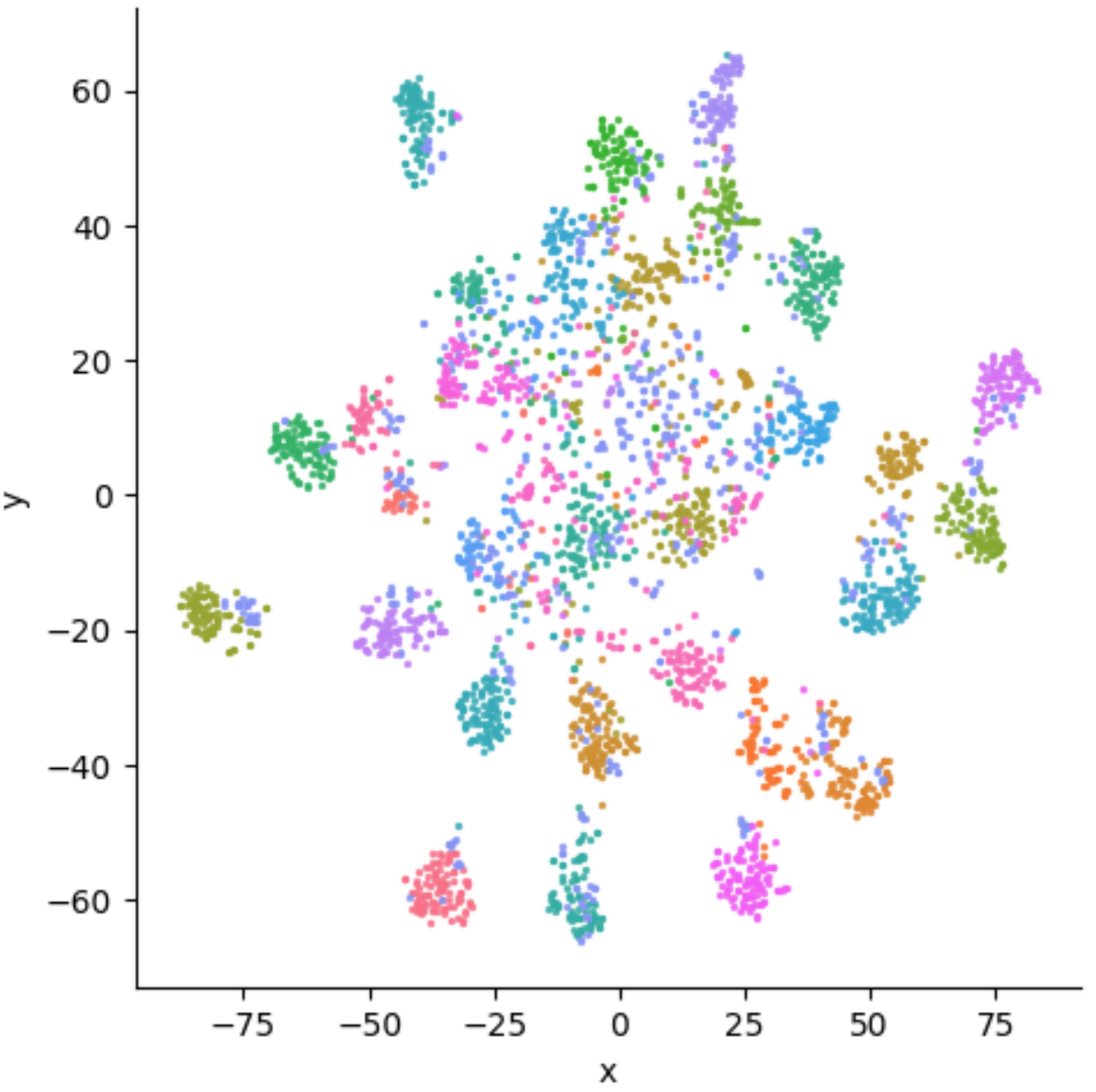}\vspace{-0.1cm}
  \caption{Amazon to DSLR}
  \label{afig:7-2}
\end{subfigure}
\begin{subfigure}{0.2\textwidth}
  \centering
  \includegraphics[width=\linewidth]{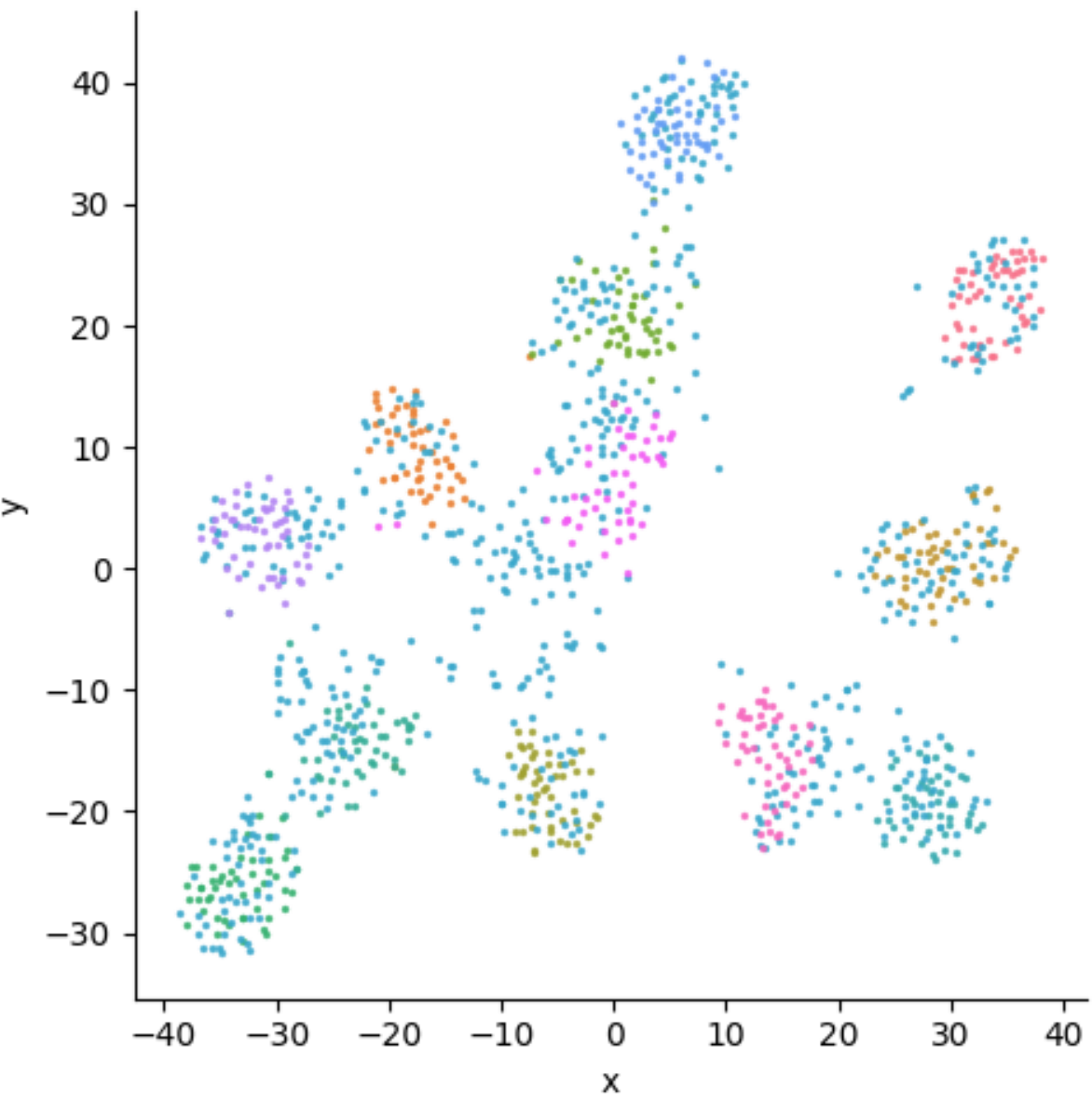}\vspace{-0.1cm}
  \caption{p to i}
  \label{afig:7-3}
\end{subfigure}
\vspace{-0.2cm}
\caption{(Best viewed in color.) Feature space in the early stages of training. \textbf{Different} from the above feature spaces, blue violet (in (a) and (b)) and deep sky blue (in (c)) denote the \textbf{target} domain and the other colors denote different classes of \textbf{source} domain.}
\label{afig:7}
\end{figure}

\begin{figure}[t]

\centering
\begin{subfigure}{0.2\textwidth}
  \centering
  \includegraphics[width=\linewidth]{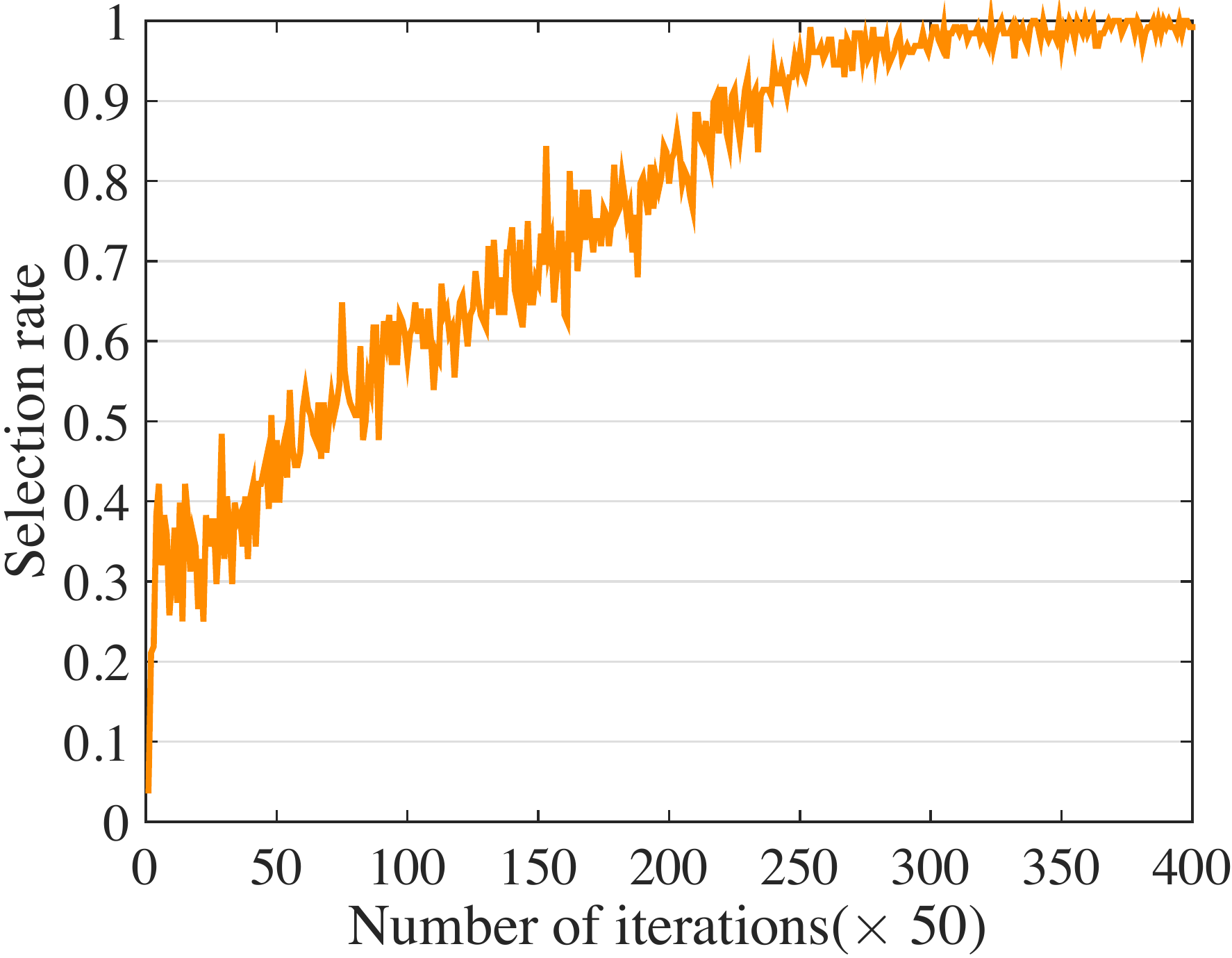}\vspace{-0.1cm}
  \caption{\emph{SVHN} to \emph{MNIST}}
  \label{afig:6-1}
\end{subfigure}
\begin{subfigure}{0.2\textwidth}
  \centering
  \includegraphics[width=\linewidth]{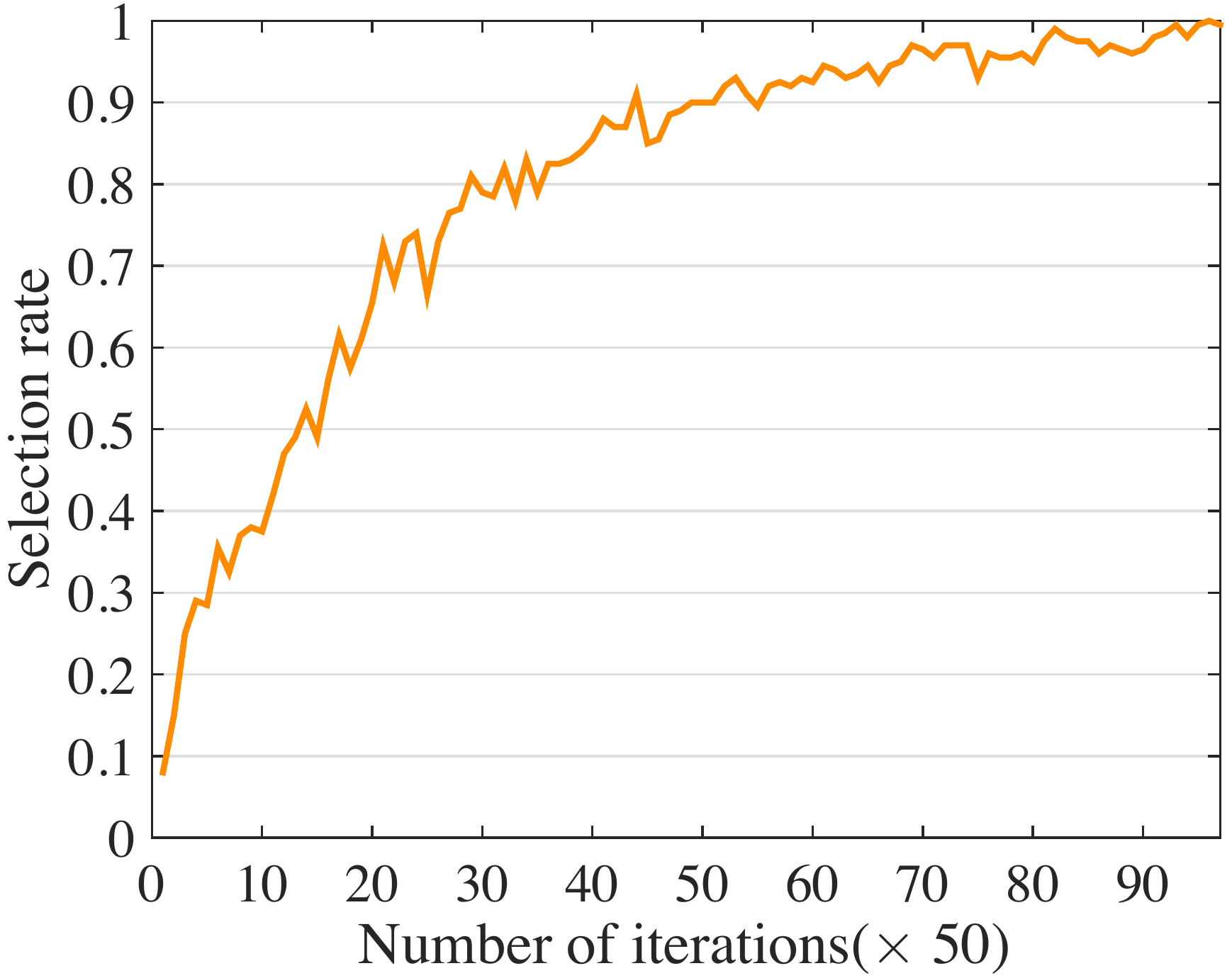}\vspace{-0.1cm}
  \caption{Amazon to DSLR}
  \label{afig:6-2}
\end{subfigure}
\begin{subfigure}{0.2\textwidth}
  \centering
  \includegraphics[width=\linewidth]{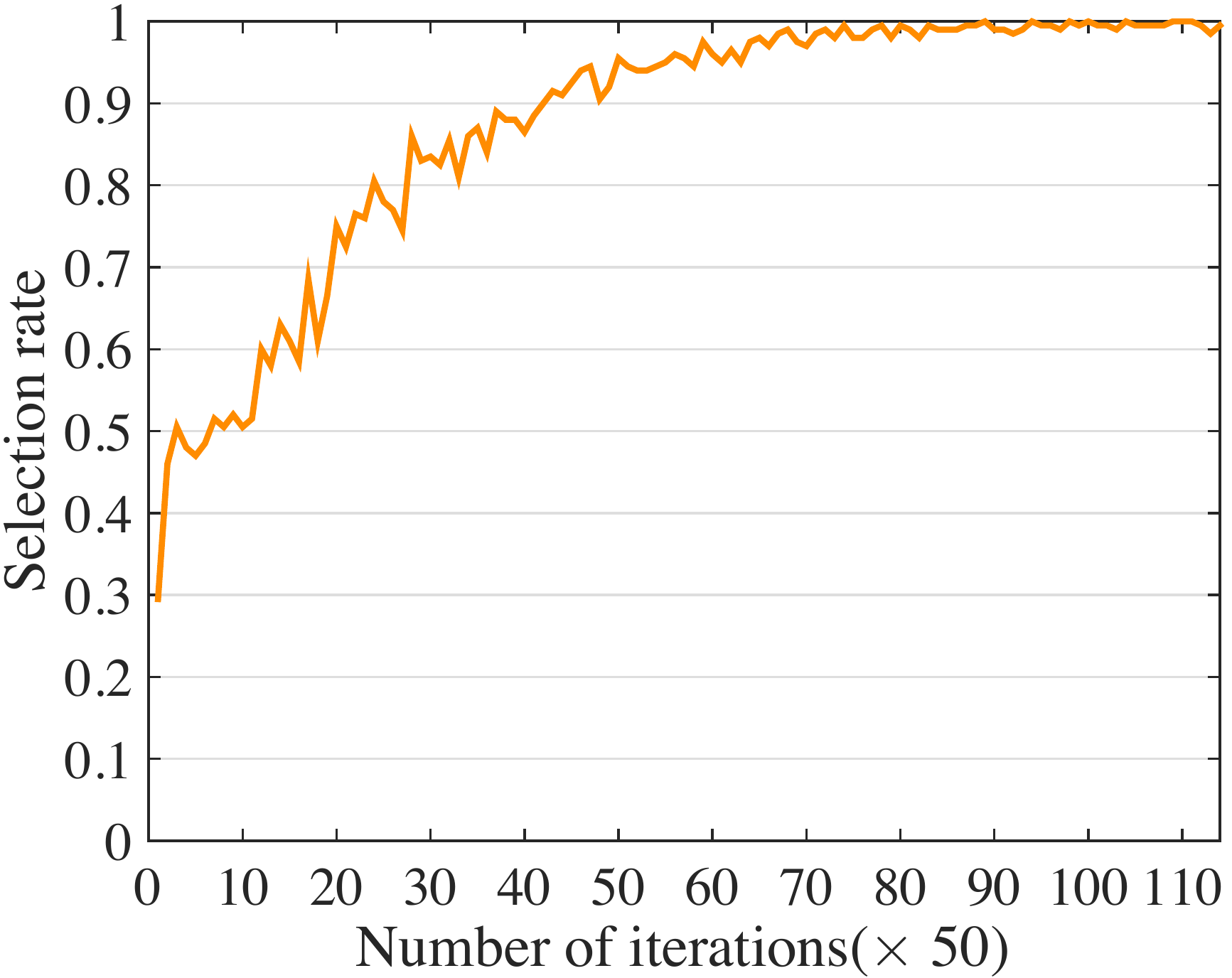}\vspace{-0.1cm}
  \caption{p to i}
  \label{afig:6-3}
\end{subfigure}
\vspace{-0.2cm}
\caption{The selection rate of the confidence-thresholding technique on different tasks.}
\label{afig:6}
\end{figure}

\section{Quantitative estimate of the divergence between domains}
When aligning the source domain and target domain via the combination of RevGrad and CAT, the loss $\mathcal{L}_d$ which is maximized w.r.t. the critic $c$ can be viewed as a lower bound of $2JSD(s, t)-2\log2$ (see [9] for the details) where $JSD$ denotes the Jensen-Shannon divergence between distributions. Therefore, we plot $\frac{1}{2}\mathcal{L}_d+\log2$ to quantitatively estimate the divergence between the two domains, following [49]. The results are shown in Fig.~\ref{afig:4} and we use the AlexNet as the classifier here. CAT can boost RevGrad significantly, leading to faster and better convergence. This group of experiments verifies that when combining CAT with the marginal distribution alignment approaches, it can provide a discriminative class-conditional alignment and bias the existing approaches to align the cluster-structure marginal distributions better.

\section{Verification of confidence-thresholding technique}
Since the source classification loss and the source discriminative clustering loss can produce strong gradients and converge quickly, the discriminative cluster structure will form in the source domain in the early stages of training. However, the classifier has not been adapted for the target domain, so a notable part of the target features will lie in the gaps between the source clusters and have low classification confidence. Therefore, the marginal alignment approaches may easily map these features into incorrect clusters, as stated in Sec. 3.2.3. To address this problem, we propose the confidence-thresholding technique which includes the fine-level structure information into marginal alignment approaches. We claim that in the training procedure, the discriminative class-conditional alignment between the two domains forms gradually, so more and more samples are going to be selected into the marginal alignment training. Here we prove these through experiments on tasks in \emph{SVHN-MNIST-USPS}, \emph{Office-31} and \emph{ImageCLEF-DA}. At first, we train the RevGrad+CAT models on the three tasks with limited iterations (\ie, 2000 iterations on \emph{SVHN} to \emph{MNIST} task, 100 iterations on Amazon to DSLR task and 100 iterations on p to i task) and plot the feature spaces of them in Fig~\ref{afig:7}. Obviously, a notable part of target samples lie in the gaps between the source clusters, especially on the \emph{SVHN} to \emph{MNIST} task which has large source and target domains. Then, we train the rRevGrad+CAT models on these tasks following the same settings, and we plot the selection rate of the confidence-thresholding technique w.r.t. the number of iterations in Fig.~\ref{afig:6}. When using this technique, we note that the selection rate monotonically increases with the number of iterations and after several thousands of iterations, the selection rate will be almost $100\%$ on the Amazon to DSLR and p to i tasks. On \emph{SVHN} to \emph{MNIST} task, we use a ramp-up function $exp(-10*(1 - min(\frac{ite-5000}{10000}, 1.)))$ as $\alpha$ after 5000 iterations, suggested by related SSL works. Therefore, after around 15000 iterations, the discriminative clustering structure forms, and then the samples are pushed far away from the decision boundaries. So almost all the samples will have confidence more than $p$ and will be selected into the domain adversarial training.

\section{Convergence}
\begin{figure}[t]
\centering
\begin{subfigure}{0.22\textwidth}
  \centering
  \includegraphics[width=\linewidth]{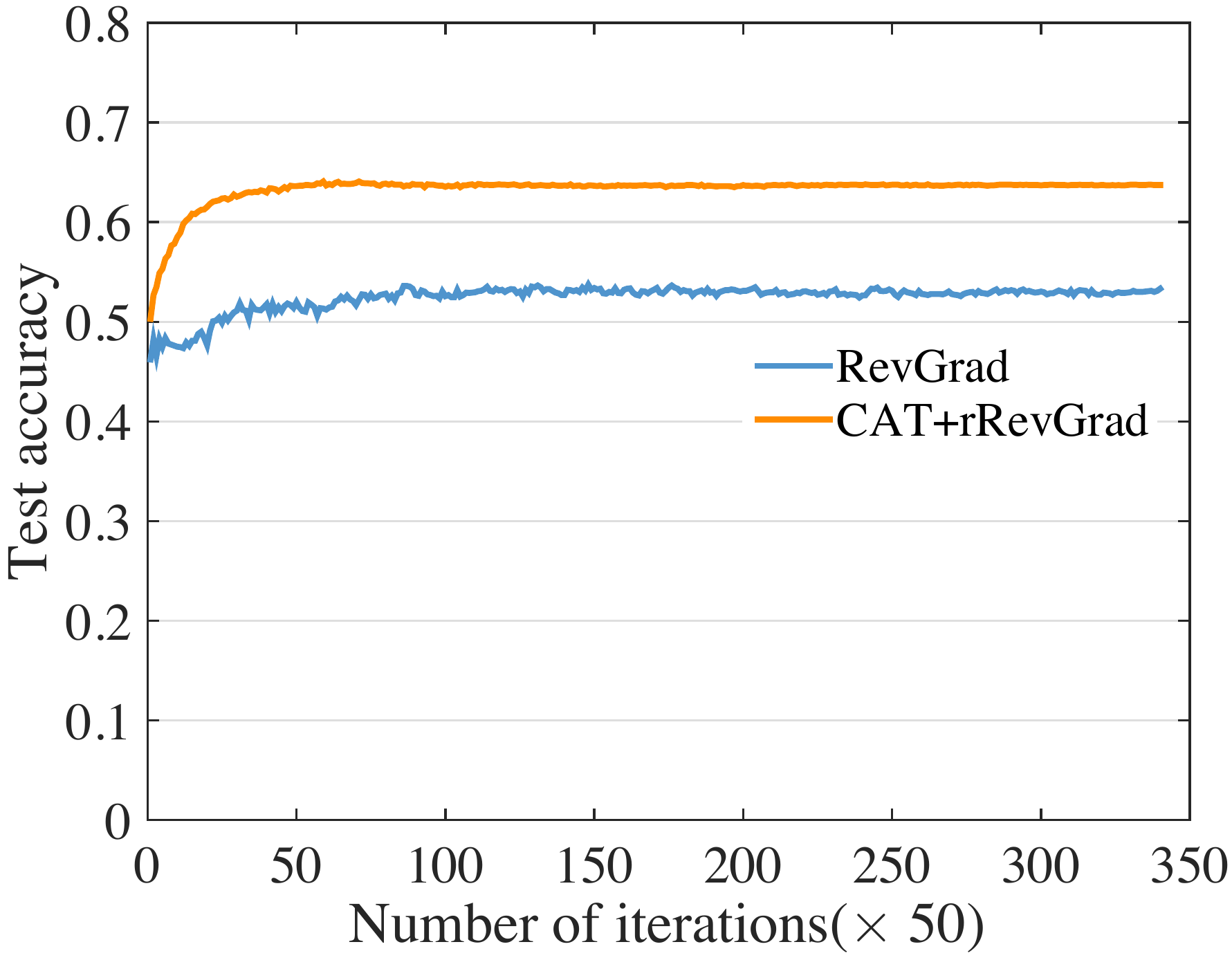}
  \caption{DSLR to Amazon}
  \label{fig:4-1}
\end{subfigure}
\begin{subfigure}{0.22\textwidth}
  \centering
  \includegraphics[width=\linewidth]{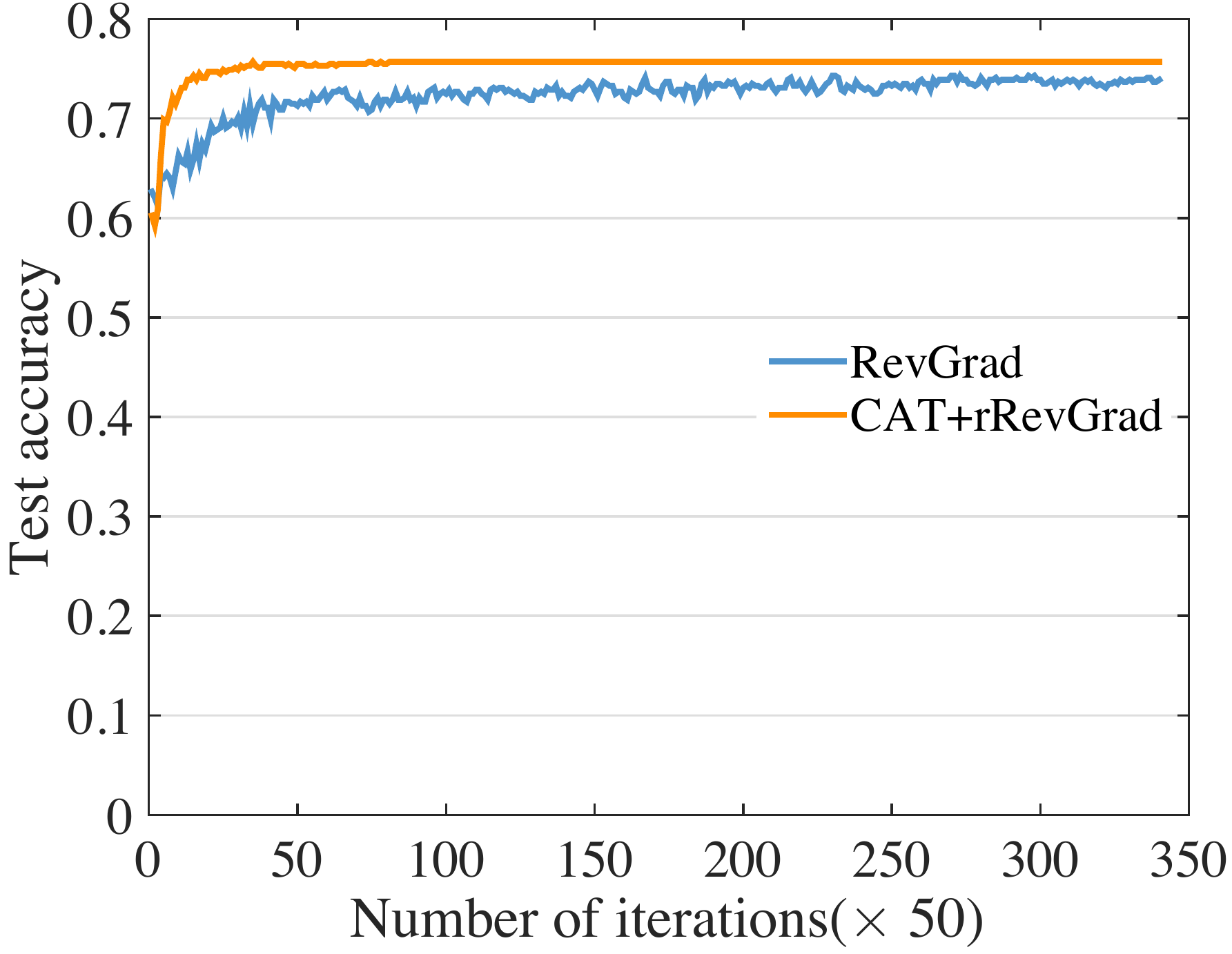}
  \caption{Amazon to DSLR}
  \label{fig:4-2}
\end{subfigure}
  \vspace{-0.3cm}
\caption{Test accuracy curves.}\vspace{-0.6cm}
\label{fig:4}
\end{figure}

To inspect how CAT converges, we plot the test accuracy with respect to the number of iterations in Fig.~\ref{fig:4}. On the two adaptation tasks using AlexNet, CAT shows similar convergence rate with RevGrad [7] but better performance.

\section{Experimental details}
On digits adaptation tasks, we use the simple LeNet with Batch Normalization after the convolutional layers and use the probability logits as features for adaptation, following [49, 44]. When combining with RevGrad [7] and rRevGrad, the critic model has a $10\rightarrow500\rightarrow500\rightarrow1$ architecture.

On more challenging tasks, we conduct experiments based on the AlexNet [15] and ResNet-50 [12] equipped with 256-D bottleneck layers after the $fc7$ and $pool5$ layers respectively (following [24, 49]). We use the features outputted by the bottleneck layers as image representations for adaptation and use a three-layer critic with $256\rightarrow1024\rightarrow1024\rightarrow1$ architecture. We finetune all the layers before the bottleneck layers in AlexNet and ResNet-50 and train the bottleneck layers and the classification layers via back propagation.

We use the stochastic gradient descent with 0.9 momentum with an annealed learning rate $\mu_p=\frac{0.01}{(1+10p)^{0.75}}$ where p changes from 0 to 1 in the training progress [7, 49] when using LeNet and AlexNet as the classifiers. The learning rate for finetuned layers is set to be the ten percent of that for layers trained from scratch. We use batches with 128 elements in experiments using LeNet, batches with 200 elements in experiments using AlexNet and batches with 36 elements in experiments using ResNet-50.

We use the same architectures and optimization settings (\eg, batch size, learning rate, optimizer and weight decay) as those of the original methods [41, 37] when combining CAT with them.

The pseudo labels are not initialized randomly. Specifically, in the first 5000 iterations, we pre-train CAT by setting $\alpha = 0$. During this, the classifier is trained to fit source data but won't overfit, thus its implicit ensemble can perform well on some target samples and provide a reliable initial set of pseudo labels. Then, we ramp-up α to activate the clustering and alignment losses to impose conditional alignment.